\documentclass{article}

\usepackage[accepted]{icml2019}
\usepackage{url}            % simple URL typesetting
\usepackage{booktabs}       % professional-quality tables
\usepackage{amsfonts}       % blackboard math symbols
\usepackage{nicefrac}       % compact symbols for 1/2, etc.
\usepackage{microtype}      % microtypography
\usepackage{lipsum}
\usepackage{booktabs} % for professional tables
\usepackage{xifthen}
\usepackage[titletoc]{appendix}
\usepackage{times}
\usepackage{amsmath, amsthm, amssymb}
\usepackage{romannum}
\usepackage{float}
\usepackage{framed}
\usepackage{mathtools}
\usepackage{thmtools,thm-restate}
\usepackage{xspace}
\usepackage{graphicx}
\usepackage{subfigure}
\usepackage[pdf]{pstricks}
\usepackage{multirow}
\usepackage{verbatim}
\usepackage{tablefootnote}
\usepackage{pifont}% http://ctan.org/pkg/pifont
\newtheorem{theorem}{Theorem}[section]
\newtheorem{lemma}[theorem]{Lemma}

\newtheorem{corollary}[theorem]{Corollary}

\newtheorem{assumption}{Assumption}

\newenvironment{definition}[1][Definition]{\begin{trivlist}
\item[\hskip \labelsep {\bfseries #1}]}{\end{trivlist}}

\newtheorem{example}{Example}
\makeatletter
\newcounter{logicase}
\newenvironment{logicase}[1][htb]
  {\renewcommand{\ALG@name}{Case}% Update algorithm name
   \let\c@algorithm\c@logicase% Update algorithm counter
   \begin{algorithm}[#1]%
  }{\end{algorithm}}
\makeatother

\newcommand{\idop}{1\!\!1}
\newcommand\defeq{\stackrel{\mathclap{\normalfont\mbox{\tiny{def}}}}{=}}

\DeclarePairedDelimiter\ceil{\lceil}{\rceil}
\DeclarePairedDelimiter\bceil{\Big\lceil}{\Big\rceil}

\DeclarePairedDelimiter\floor{\lfloor}{\rfloor}
\DeclarePairedDelimiter\bfloor{\Big\lfloor}{\Big\rfloor}

\DeclarePairedDelimiter\brackets{[}{]}
\newcommand{\Ex}[1]{\E\brackets*{#1}}

\newcommand{\TOPRHO}{{\mathcal{TOP}_\rho}}
\newcommand{\A}{\mathcal{A}}

\newcommand{\F}{\mathcal{F}}

\newcommand{\I}{\mathcal{I}}

\newcommand*{\hatp}{\hat{p}}
\newcommand{\SUBSET}{\textsc{Subset}\@\xspace}
\newcommand{\QF}{\textsc{Q-F}\@\xspace}

\newcommand{\QFK}{$(k, m, n)$\@\xspace}

\newcommand{\QP}{\textsc{Q-P}\@\xspace}
\newcommand{\QPK}{$(k, \rho)$\@\xspace}

\newcommand{\FF}{$\mathcal{F}_2$\@\xspace}
\newcommand{\KQP}{\textsc{KQP}-1\@\xspace}

\makeatletter
\newcommand*{\PP}{%
    \@ifnextchar{.}%
        {$\mathcal{P}_2$}%
        {$\mathcal{P}_2$\@\xspace}%
}
\makeatother
\makeatletter
\newcommand*{\PPP}{%
    \@ifnextchar{.}%
        {$\mathcal{P}_3$}%
        {$\mathcal{P}_3$\@\xspace}%
}
\makeatother
\makeatletter
\newcommand{\LUCB}{%
    \@ifnextchar{.}%
        {\textsc{LUCB1}}%	
        {\textsc{LUCB1}\@\xspace}%
}
\makeatother
\makeatletter
\newcommand*{\HALVING}{%
    \@ifnextchar{.}%
        {\textsc{Halving}}%	
        {\textsc{Halving}\@\xspace}%
}
\makeatother
\makeatletter
\newcommand*{\GHALVING}{%
    \@ifnextchar{.}%
        {\textsc{Halving}-k-m}%	
        {\textsc{Halving}-k-m\@\xspace}%
}
\makeatother

\newcommand{\GLUCB}{\textsc{LUCB}-k-m\@\xspace}

\DeclareMathOperator*{\E}{\mathop{\mathbb{E}}}

\newcommand*{\st}{such that\@\xspace}

\newcommand*{\iid}{i.i.d.\@\xspace}
\makeatletter
\newcommand*{\iispa}{%
    \@ifnextchar{.}%
        {\emph{p.i.s.p.a}}%	
        {\emph{p.i.s.p.a.}\@\xspace}%
}
\makeatother
\makeatletter
\newcommand*{\etc}{%
    \@ifnextchar{.}%
        {etc}%	
        {etc.\@\xspace}%
}
\makeatother

\makeatletter
\newcommand{\ALOOP}[1]{\ALC@it\algorithmicloop\ #1%
  \begin{ALC@loop}}
\newcommand{\ENDALOOP}{\end{ALC@loop}\ALC@it\algorithmicendloop}
\renewcommand{\algorithmicrequire}{\textbf{Input:}}
\renewcommand{\algorithmicensure}{\textbf{Output:}}

\makeatother

\begin{document}
% \maketitle
\twocolumn[
\icmltitle{PAC Identification of Many Good Arms in Stochastic Multi-Armed Bandits}
\icmlsetsymbol{equal}{*}
\begin{icmlauthorlist}
\icmlauthor{Arghya Roy Chaudhuri}{iitb}
\icmlauthor{Shivaram Kalyanakrishnan}{iitb}
\end{icmlauthorlist}

\icmlaffiliation{iitb}{Department of Computer Science and Engineering, Indian Institute of Technology Bombay, Mumbai 400076, India}

\icmlcorrespondingauthor{Arghya Roy Chaudhuri}{arghya@cse.iitb.ac.in}
\icmlcorrespondingauthor{Shivaram Kalyanakrishnan}{shivaram@cse.iitb.ac.in}

% You may provide any keywords that you
% find helpful for describing your paper; these are used to populate
% the "keywords" metadata in the PDF but will not be shown in the document
% \icmlkeywords{Multi-Armed Bandit, Bounded Arm Memory, Regret Minimisation}
% \vskip 0.3in
]
\printAffiliationsAndNotice{}
\begin{abstract}
We consider the problem of identifying any $k$ out of the best $m$ arms in an $n$-armed stochastic multi-armed bandit. Framed in the PAC setting, this particular problem
generalises both the problem of ``best subset selection''~\cite{bib:explorem} and that of
selecting ``one out of the best m'' arms~\cite{bib:arcsk2017}. In applications such as crowd-sourcing and drug-designing, identifying a single good solution is often not sufficient. Moreover,
finding the best subset might be hard due to the presence of many indistinguishably close solutions. Our generalisation of identifying exactly $k$ arms out of the best $m$, where $1 \leq k \leq m$, serves as a more effective alternative.
We present a lower bound on the worst-case sample complexity for general $k$, and
a fully sequential PAC algorithm, \GLUCB, which is more sample-efficient on easy instances.
Also, extending our analysis to infinite-armed bandits, we present a PAC algorithm that is independent of $n$, 
which identifies an arm from the best $\rho$ fraction of arms using at most an additive poly-log number of samples than
compared to the lower bound, 
thereby improving over~\citet{bib:arcsk2017} and \citet{Aziz+AKA:2018}. 
The problem of identifying $k > 1$ distinct arms from the best $\rho$ fraction is not always well-defined; 
for a special class of this problem, we present lower and upper bounds. Finally, through a reduction, we establish a relation between upper bounds for the ``one out of the best $\rho$'' problem for infinite instances and the ``one out of the best $m$'' problem for finite instances. We conjecture that it is more efficient to solve ``small" finite instances using the latter formulation, rather than going through the former.
\end{abstract}
\section{Introduction}
\label{sec:intro}
The stochastic multi-armed bandit~\citep{Robbins:1952,bib:mabdefbook} is
a well-studied abstraction of decision making under uncertainty. Each \emph{arm} of a bandit represents a decision. A \emph{pull} of an arm represents taking the associated decision, which produces a real-valued reward. The reward is drawn \iid from a distribution corresponding to the arm, independent of the pulls of other arms. At each round, the experimenter may consult the preceding history of pulls and rewards to decide which arm to pull.

The traditional objective of the experimenter is to maximise the expected cumulative reward over a horizon of pulls, or equivalently, to minimise the \textit{regret} with respect to always pulling an optimal arm. Achieving this objective requires a careful balance between \textit{exploring}  (to reduce uncertainty about the arms' expected rewards) and \textit{exploiting} (accruing high rewards). Regret-minimisation algorithms have been used in a variety of applications, including clinical trials~\citep{Robbins:1952}, adaptive routing~\citep{Awerbuch+K:2008}, and recommender systems~\citep{Li+CLS:2010}.

Of separate interest is the problem of \textit{identifying} an arm with the highest mean reward~\citep{Bechhofer:1958,Paulson:1964,bib:evendar1}, under what is called the ``pure exploration'' regime. For applications such as product testing~\citep{Audibert+BM:2010} and strategy selection~\citep{bib:sergiu}, there is a dedicated phase in the experiment in which the rewards obtained are inconsequential. Rather, the objective is to identify the best arm either (1) in the minimum number of trials, for a given confidence threshold~\citep{bib:evendar1,bib:explorem}, or alternatively, (2) with minimum error, after a given number of trials~\citep{Audibert+BM:2010,Carpentier+V:2015}. Our investigation falls into the first category, which is termed the ``fixed confidence'' setting. Conceived by Bechhofer (\citeyear{Bechhofer:1958}), 
best-arm-identification in the fixed confidence setting has received a significant amount of attention over the years~\citep{bib:evendar1,Gabillon+GLB:2011,Karnin+KS:2013,Jamieson+MNB:2014}.
The problem has also been generalised to identify the best subset of
arms~\citep{bib:explorem,bib:lucb}.

More recently, \citet{bib:arcsk2017} have introduced the probem of identifying a single arm from among the best $m$ in an $n$-armed bandit. This formulation is particularly useful when the number of arms is large, and in fact is a viable alternative even when the number of arms is \textit{infinite}. In many practical scenarios, however, it is required to identify more than a single good arm. For example, imagine that a company needs to complete a task that is too large to be accomplished by a single worker, 
but which can be broken into $5$ subtasks, each capable 
of being completed by one worker. Suppose there are a total of $1000$ workers, and an indepdendent pilot survey has revealed that at least $15\%$ of them have the skills to complete the subtask. To address the company's need, surely it would \textit{suffice} to identify the best $5$ workers for the subtask. However, if workers are to be identified based on a skill test that has stochastic outcomes, it would be unnecessarily expensive to indeed identify the ``best subset''. Rather, it would be enough to merely identify any $5$ workers from among the best $150$. This is precisely the problem we consider in our paper:
identifying any $k$ out of the best $m$ arms in an $n$-armed bandit. In addition to distributed crowdsourcing~\citep{Tran-Thanh+SRJ:2014}, applications of this problem include the management of large sensor networks~\citep{Mousavi+HHD:2016}, wherein multiple 
reliable sensors must be identified using minimal testing, and in drug design~\citep[Chapter 43]{McDuffie+OJ:2016}, to identify a promising set of candidate biomarkers. 

The problem assumes equal significance from a theoretical standpoint, since it generalises both the ``best subset selection'' problem~\citep{bib:explorem} (taking $k = m$) and that of selecting a ``single arm from the best subset''~\citep{bib:arcsk2017} (taking $k = 1$). Unlike best subset selection, the problem remains feasible to solve even when $n$ is large or infinite, as long as $m/n$ is some constant $\rho > 0$. Traditionally, infinite-armed bandits have been tackled by resorting to side information such as distances between arms~\citep{bib:agracntregret,bib:kleincntregret} or the structure of their distribution of rewards~\citep{bib:wang2008}. This approach introduces additional parameters, which might not be easy to tune in practice. Alternatively, good arms can be reached merely by selecting arms \textit{at random} and testing them by pulling. This latter approach has been applied successfully both in the regret-minimisation setting~\citep{Herschkorn+PR:1996} and in the fixed-confidence setting~\citep{bib:sergiu,bib:arcsk2017}. Our formulation paves the way for identifying ``many'' ($k$) ``good'' (in the top $m$ among $n$) arms in this manner. 

%We provide lower and upper bounds on the worst case sample complexity of our ``$(k, m, n)$'' problem, which match up to a constant factor. While generalising existing lower bounds for $(1, m, n)$ and $(m, m, n)$, and also the upper bounds for $(m, m, n)$, we improve upon the previous upper bound for $(1, m, n)$~\citep{bib:arcsk2017} by eliminating a logarithmic factor. We furnish a fully sequential algorithm for $(k, m, n)$, which generalises the LUCB algorithm for $(m, m, n)$~\citep{bib:lucb}, and in fact empirically outperforms the $\F_2$ algorithm for $(1, m, n)$~\citep{bib:arcsk2017}. We extend our algorithms to the infinite setting, where our lower and upper bounds match up to a factor of $O(\log(k))$. PARAGRAPH NEEDS CHECKING.

The interested reader may refer to Table~\ref{tab:prevressummary} right away for a summary of our theoretical results, which are explained in detail after formally specifying the $(k, m, n)$ and $(k, \rho)$ problems in Section~\ref{sec:problemdefinitionandcontributions}. In Section~\ref{sec:finiteinst} we present our algorithms and analysis for the finite setting, and in Section~\ref{sec:infinitemab} for the infinite setting. We present experimental results in Section~\ref{sec:expt}, and conclude with a discussion in Section~\ref{sec:conclusion}.

% Please add the following required packages to your document preamble:
% \usepackage{multirow}
\begin{table*}[t]
\centering
\caption{Lower and upper bounds on the expected sample complexity (worst case over problem instances). The bounds for \QPK, $k > 1$ are for the special class of ``at most $k$-equiprobable" instances.
}
\label{tab:prevressummary}
\resizebox{1\textwidth}{!}{
\begin{tabular}{l|cc|c}
\hline
\multicolumn{1}{c|}{Problem} & Lower Bound & Previous Upper Bound & Current Upper Bound \\ \hline
\multirow{2}{*}{\begin{tabular}[c]{@{}l@{}}$(1, 1, n)$\\ Best-Arm \end{tabular}} & $\Omega\left(\frac{n}{\epsilon^2}\log\frac{1}{\delta}\right)$ & $O\left(\frac{n}{\epsilon^2}\log\frac{1}{\delta}\right)$ & \multirow{2}{*}{Same as previous} \\
 & {\footnotesize \citep{bib:mannor2004}} & {\footnotesize \citep{bib:evendar1}} &  \\ \hline
\multirow{2}{*}{\begin{tabular}[c]{@{}l@{}}$(m, m, n)$\\ \textsc{Subset}\end{tabular}} & $\Omega\left(\frac{n}{\epsilon^2}\log\frac{m}{\delta}\right)$ & $O\left(\frac{n}{\epsilon^2}\log\frac{m}{\delta}\right)$ & \multirow{2}{*}{Same as previous} \\
 & {\footnotesize \citep{bib:lucb}} & {\footnotesize \citep{bib:explorem}} &  \\ \hline
\multirow{2}{*}{\begin{tabular}[c]{@{}l@{}}$(1, m, n)$\\ \textsc{Q-F}\end{tabular}} & $\Omega\left(\frac{n}{m\epsilon^2}\log\frac{1}{\delta}\right)$ & $O\left(\frac{n}{m\epsilon^2}\log^{2}\frac{1}{\delta}\right)$ & $O\left(\frac{1}{\epsilon^2}\left(\frac{n}{m}\log\frac{1}{\delta} + \log^2\frac{1}{\delta}\right)\right)$ \\
 & \multicolumn{2}{c|}{{\footnotesize \citep{bib:arcsk2017}}} & {\footnotesize \textbf{This paper}} \\ \hline
\multirow{2}{*}{\begin{tabular}[c]{@{}l@{}}$(k, m, n)$\\ $\textsc{Q-F}_k$\end{tabular}} & $\Omega\left(\frac{n}{(m-k+1)\epsilon^2}\log\frac{\binom{m}{k-1}}{\delta}\right)$ & - & $O\left(\frac{k}{\epsilon^2}\left(\frac{n\log k}{m}\log\frac{k}{\delta} + \log^2\frac{k}{\delta}\right)\right)^*$ \\
 & {\footnotesize \textbf{This paper}} &  & {\footnotesize \textbf{This paper}\; (*\text{for}\; $k \geq 2$}) \\ \hline
\multirow{2}{*}{\begin{tabular}[c]{@{}l@{}}$(1, \rho)$ ($|\A| = \infty$)\\ \textsc{Q-P}\end{tabular}} & $\Omega\left(\frac{1}{\rho\epsilon^2}\log\frac{1}{\delta}\right)$ & $O\left(\frac{1}{\rho\epsilon^2}\log^{2}\frac{1}{\delta}\right)$ & $O\left(\frac{1}{\epsilon^2}\left(\frac{1}{\rho}\log\frac{1}{\delta} + \log^2\frac{1}{\delta}\right)\right)$ \\
 & \multicolumn{2}{c|}{{\footnotesize \citep{bib:arcsk2017}}} & {\footnotesize \textbf{This paper}} \\ \hline
\multirow{2}{*}{\begin{tabular}[c]{@{}l@{}}$(k, \rho)$ ($|\A| = \infty$)\\ $\textsc{Q-P}_k$  \end{tabular}} & $\Omega\left(\frac{k}{\rho\epsilon^2}\log\frac{k}{\delta}\right)$ & - & $O\left(\frac{k}{\epsilon^2}\left(\frac{\log k}{\rho}\log\frac{k}{\delta} + \log^2\frac{k}{\delta}\right)\right)^*$\\
 & {\footnotesize \textbf{This paper}} &  &  {\footnotesize \textbf{This paper}  (*for a special class with\; $k \geq 2$)}\\ \hline
\end{tabular}
}
\end{table*}
\section{Problem Definition and Contributions}
\label{sec:problemdefinitionandcontributions}

Let $\A$ be the set of arms in our given bandit instance. With each arm $a \in \A$, there is
an associated reward distribution supported on a subset of $[0, 1]$, with mean $\mu_a$. When pulled, arm $a \in \A$ produces a reward drawn \iid from the corresponding distribution, and independent of the pulls of other arms.
%Without the loss of generality, we can assume, for any two arms $a_i, a_j \in \A$, $\mu_{a_i} \geq \mu_{a_j}$, whenever $i \leq j$. The experimenter does not have any knowledge regarding reward distributions associated with the arms, except the fact that, the rewards are \iid samples from the corresponding reward distribution bounded in $[0, 1]$.
At each round, based on the preceding sequence of pulls and rewards, an algorithm either decides
which arm to pull, or stops and returns a set of arms.

For a finite bandit instance with $n$ arms, we take $\A = \{a_{1}, a_{2}, \dots, a_{n}\}$, and assume, without
loss of generality, that for arms $a_i, a_j \in \A$, $\mu_{a_i} \geq \mu_{a_j}$ whenever $i \leq j$. Given a tolerance $\epsilon \in [0, 1]$ and  $m \in \{1, 2,  \dots, n\}$, we call an arm $a \in \A$
 $(\epsilon, m)$-optimal if $\mu_a \geq \mu_{a_m} - \epsilon$. We denote the set of 
all the $(\epsilon, m)$-optimal arms as  $\mathcal{TOP}_m(\epsilon) \defeq \{a: \mu_a \geq \mu_{a_m} -\epsilon\}$. For brevity we denote $\mathcal{TOP}_m(0)$ as $\mathcal{TOP}_m$. 

\begin{definition}{\QFK Problem.}
An instance of the \QFK problem is of the form $(\A, n, m, k, \epsilon, \delta)$, where $\A$ is a set of arms with $|\A| = n \geq 2$; $m \in \{1, 2, \dots, n - 1\}$; $k \in \{1, \dots, m\}$; tolerance $\epsilon \in (0, 1]$; and mistake probability $\delta \in (0,1]$. An algorithm $\mathcal{L}$ is said to solve
\QFK if for every instance of \QFK, it terminates
with probability 1, and returns $k$ \emph{distinct} $(\epsilon, m)$-optimal arms with probability at least $1-\delta$.
\end{definition}

The \QFK problem is interesting from a theoretical standpoint because it covers an entire range of problems,
with single-arm identification ($m = 1$) at one extreme and subset selection ($k = m$) at the other. Thus, any bounds on the sample
complexity of \QFK also apply to \QF~\citep{bib:arcsk2017} and to \textsc{Subset}~\citep{bib:explorem}. 
In this paper, we show that any algorithm that solves \QFK must incur 
$\Omega\left(\frac{n}{(m - k + 1)\epsilon^{2}} \log\left(\frac{\binom{m}{k - 1}}{\delta}\right)\right)$ 
pulls for some instance of the problem. We are unaware of bounds in the fixed-confidence setting that 
involve such a combinatorial term inside the logarithm. 
% Interestingly, we are able to show that this dependence is optimal: we do so by furnishing an algorithm that achieves a sample complexity within a constant factor of the lower bound.

Table~\ref{tab:prevressummary} places our bounds in the context of previous results. 
The bounds shown in the table consider the worst-case across problem instances; 
in practice one can hope to do better on easier problem instances by adopting a fully 
sequential sampling strategy. Indeed we adapt the \textsc{LUCB1} algorithm~\citep{bib:lucb} 
to solve \QFK, denoting the new algorithm \GLUCB. Our analysis shows that for $k=1$, and $k=m$, 
the upper bound on the sample complexity of this algorithm matches with those of 
$\F_2$~\citep{bib:arcsk2017} and \LUCB~\citep{bib:lucb}, respectively, up to a multiplicative constant.
Empirically, \GLUCB with $k = 1$ appears to be more efficient than $\F_2$ for solving \QF.

Along the same lines that \citeauthor{bib:arcsk2017} (\citeyear{bib:arcsk2017}) define the \QP problem for infinite instances, we define a generalisation of \QP for selecting many good arms, which we denote \QPK. Given a set of arms $\A$, a sampling
distribution $P_\A$, $\epsilon \in (0,1]$, and $\rho \in [0, 1]$, an arm $a \in \A$
is called $[\epsilon, \rho]$-optimal if
$P_{a' \sim P_\A} \{\mu_a \geq \mu_{a'} -\epsilon\} \geq 1 - \rho$. For $\rho, \epsilon \in [0,1]$, we define the set of all $[\epsilon, \rho]$-optimal arms as
$\TOPRHO(\epsilon)$. As before, we denote $\TOPRHO(0)$ as $\TOPRHO$.
%Also, let  $\mu_\rho$ be the $(1-\rho)$-th quantile of the probability distribution over arm means of the arms induced by $P_\A$, and hence, for all $a \in \TOPRHO$, and $a' \in \A \setminus \TOPRHO$, $\mu_{a'} < \mu_\rho \leq \mu_a$.
A straightforward generalisation of \QP is as follows.
% \begin{definition}{(\QPK)}
% An instance of \QPK is fixed by a bandit instance
% with a set of arms $\A$; a probability distribution over $P_\A$ over $\A$;
%  $\rho \in (0, 1]$; tolerance $\epsilon \in (0, 1]$; and mistake probability $\delta \in (0,1]$.
% An algorithm $\mathcal{L}$ is said to solve
% \QPK if for every instance of \QPK as input, $\mathcal{L}$ terminates
% with probability 1, and returns $k$ distinct $[\epsilon, \rho]$-optimal arms with probability
% at least $1-\delta$.
% \end{definition}

\begin{definition}{\QPK Problem.}
An instance of the \QPK problem
is of the form $(\A, P_\A, k, \rho, \epsilon, \delta)$, where $\A$ is a set of arms; $P_\A$ is a probability distribution 
over $\A$; quantile fraction $\rho \in (0, 1]$; tolerance $\epsilon \in (0, 1]$; and mistake probability $\delta \in (0,1]$. Such an instance is \emph{valid} if $|\TOPRHO| \geq k$, and \textit{invalid} otherwise.
 % Given a valid instance of \QPK, 
An algorithm $\mathcal{L}$ is said to solve
\QPK, if for every \emph{valid} instance of \QPK, $\mathcal{L}$ terminates
with probability 1, and returns $k$ \emph{distinct} $[\epsilon, \rho]$-optimal arms with probability
at least $1-\delta$.
\end{definition}

\paragraph{At most $k$-equiprobable instances.} Observe that \QPK is well-defined only if the given instance has at least $k$ distinct arms in $\TOPRHO$; we term such an instance
\textit{valid}. It is worth noting that even valid instances can require an arbitrary amount of computation
to solve. For example, consider an instance with $k > 1$ arms in $\TOPRHO$, one among which has a probability $\gamma$ of being picked by $P_{\A}$, and the rest each a probability of $(\rho - \gamma) / (k - 1)$. Since the arms have to be identified by sampling from $P_{\A}$, the probability of identifying the latter $k - 1$ arms diminishes to $0$ as $\gamma \to \rho$, calling for an infinite number of guesses. To avoid this scenario, we restrict our analysis to a special class of valid instances in which $P_{\A}$ allocates no more than $\rho/k$ probability to any arm in 
$\TOPRHO$. We refer to such instances as ``at most $k$-equiprobable'' instances. Formally,
a \QPK problem instance given by $(\A, P_\A, k, \rho, \epsilon, \delta)$ is called
``at most $k$-equiprobable'' if 
$\forall a \in \TOPRHO$, $\Pr_{\mathbf{a}' \sim P_\A}\{\mathbf{a}' = a\} \leq \frac{\rho}{k}$.\footnote{In a recent paper, \citet{Ren+LS:2018} claim to solve the \QPK problem. However, they do not notice that the problem can be ill-posed. Also, even with an at most $k$-equiprobable instance as input, their algorithm fails to escape the $(1/\rho)\log^2(1/\delta)$ dependence.}

Note that any instance of the $(1, \rho)$ or \QP~\citep{bib:arcsk2017} problem is necessarily valid and at most $1$-equiprobable. Interestingly, we improve upon the existing  upper bound for this problem, so it matches the lower bound up to an \textit{additive} $O\left(\frac{1}{\epsilon^{2}}\log^2\frac{1}{\delta}\right)$ term. Below we summarise our contributions.
% Previous upper bounds for this problem vary as $\log^{2}(1/\delta)$~\citep{Aziz+AKA:2018}. 

% \filler{For ``at most $k$-equiprobable'' instances of \QPK, we show a lower bound of $\Omega\left(\frac{k}{\rho\epsilon^2}\log\left(\frac{k}{\delta}\right)\right)$. We also present an algorithm that solves \QPK, and whose sample complexity on ``at most $k$-equiprobable'' instances is tight up to a factor of $O(\log(k))$.}
\begin{enumerate}
    \item We generalise two previous problems---\QF and \textsc{Subset}~\citep{bib:arcsk2017}---via \QFK. In Section~\ref{sec:finiteinst} we derive a lower bound on the worst case sample 
    complexity to solve \QFK, which generalises existing lower bounds for \QF and \SUBSET.
    
    \item  In Section~\ref{sec:adaptive} we extend \textsc{LUCB1}~\citep{bib:lucb} 
    to present a fully-sequential algorithm---\emph{\underbar{LUCB} for $\underbar{k}$ out of $\underbar{m}$}
    or \GLUCB---to solve \QFK. 
    We shows that for $k=1$, and $k=m$ the upper bound on its expected sample complexity
    matches with those of $\F_2$, and \LUCB, respectively, up to a constant factor.
    
    \item  In Section~\ref{sec:infinitemab} we present algorithm \PPP to solve 
    \QP with a sample complexity that is an additive $O((1/\epsilon^{2})\log^2(1/\delta))$ term away from the lower bound.
    We extend it to an algorithm \KQP for solving at most $k$-equiprobable \QPK instances. Also, \PPP and \KQP can
    solve \QF and \QFK respectively, and their sample complexities are the tightest 
    instance-independent upper bounds as yet.

    \item In Section~\ref{subsec:qpreducub} we present a general relation between the 
    upper bound on the sample complexities for solving \QF and \QP. This helps in effectively transferring
    any improvement in the upper bound on the former to the latter. Also, we conjecture
    the existence of a class of \QF instances that can be solved more efficiently than their 
    ``corresponding" \QP instances.
    % show that how improving the 
    % that if there exists an algorithm that solves
    % \QF with sample complexity within a constant factor its lower bound, then there exists an algorithm that
    % solves \QP with a sample complexity within a constant factor of the latter's lower bound.

    \item In Section~\ref{sec:expt} we experimentally show that \GLUCB is
    significantly more efficient than $\F_2$ for solving \QF.
    %, and justify the result.
    
\end{enumerate}

\section{Algorithms for Finite Instances}
\label{sec:finiteinst}

We begin our technical presentation by furnishing a lower bound on the sample complexity of algorithms for \QFK.

\subsection{Lower Bound on the Sample-Complexity}
\label{subsec:lbsckmn}

\begin{restatable}{theorem}{thmlbmainthm}[Lower Bound for \QFK]
\label{thm:lbmainthm}
Let $\mathcal{L}$ be an algorithm that solves \QFK. Then, there exists an 
instance $(\A, n, m, k, \epsilon, \delta)$, 
with $0< \epsilon \leq \frac{1}{\sqrt{32}}$, $0 < \delta \leq \frac{e^{-1}}{4}$, 
and $n \geq 2m$, $1 \leq k \leq m$, on which the expected number of pulls 
performed by $\mathcal{L}$ is at least $\frac{1}{18375}. \frac{1}{\epsilon^2}. \frac{n}{m-k+1}\ln\frac{\binom{m}{k - 1}}{4\delta}$.
\end{restatable}

The detailed proof of the theorem is given in Appendix~\ref{app:lowerboundqfk}. 
The proof generalises lower bound proofs for both $(m, m, n)$
~\citep[see Theorem 8]{bib:lucb} and $(1, m, n)$~\citep[see Theorem 3.3]{bib:arcsk2017}.
The core idea in these proofs is to consider two sets of bandit instances,
$\mathcal{I}$ and $\mathcal{I}^{\prime}$, such that over ``short'' trajectories, 
an instance from $\mathcal{I}$ will yield the same reward sequences as a corresponding 
instance from $\mathcal{I}^{\prime}$, with high probability. Thus, any algorithm 
will return the same set of arms for both instances, with high probability. 
However, by construction, no set of arms can be simultaneously correct for both 
instances---implying that a correct algorithm must encounter sufficiently ``long'' 
trajectories. Our main contribution is in the design of 
$\mathcal{I}$ and $\mathcal{I}^{\prime}$ when $k \in \{1, 2, \dots, m\}$ 
(rather than exactly $1$ or $m$) arms have to be returned.

Our algorithms to achieve improved \textit{upper} bounds for \QF and \QFK 
(across bandit instances) follow directly from methods we design for the 
infinite-armed setting in Section~\ref{sec:infinitemab} (see Corollary~\ref{cor:qffromqptighter} 
and Corollary~\ref{cor:qfkfromqpktighter}). In the remainder of this section, we present a fully-sequential algorithm for \QFK whose expected sample complexity varies with the ``hardness'' of the input instance.

\subsection{An Adaptive Algorithm for Solving \protect\QFK} %\textsc{LUCB}$(k, m)$
\label{sec:adaptive}

% Although there exist instances of \QFK that require within a constant factor of the sample complexity of \GHALVING, the algorithm is likely to be wasteful on ``easy'' instances, in which arms in $\TOPM$ are well-separated from the remaining arms.

%We present an adaptive algorithm, \GLUCB, for solving $(k,m,n)$ and analyse its sample complexity.

Algorithm~\ref{alg:glucb} describes \GLUCB, a fully sequential algorithm, which for $k=1$ has the same
guarantee on sample-complexity as \FF, but empirically appears to be more economical. The algorithm  generalises \LUCB~\cite{bib:lucb}, which solves $(m, m, n)$. 

% However, it differs from
% \LUCB in a subtle manner, due to the very definition of the problem. Like \QF, \QFK
% assumes multiple solutions if $k < m$. On the other hand, it differs from \QF as it
% can solve \SUBSET of size $m \geq 1$ for $k=m$.

% \begin{algorithm}[]
% \small{
% \caption{\GLUCB: Algorithm to select $k$ $(\epsilon, m)$-optimal arms}
% \label{alg:glucb}
% \DontPrintSemicolon% instead
%  \KwIn{$\mathcal{A}$ (\st $|\mathcal{A}| = n$), $k, m, \epsilon, \delta$.}
%  \KwOut{$k$ distinct $(\epsilon,m)$-optimal arms from $\mathcal{A}$.}
%  Pull each arm $a  \in \mathcal{A}$ once. Set $t = n$.\;
%  \Do{$ ucb({l_*^t}, t+1) - lcb({h_*^t}, t+1) > \epsilon.$} { \label{ln:stpkoutofm}
%      $t = t + 1$.\;
% %      $A_1^t = \{a : a' \in \mathcal{A}, \hatp_a = \hatp_{a'}\}$ \st $|A_1^t| = k$.\;
% %      $A_3^t = \{a : a' \in \mathcal{A}, \hatp_a = \hatp_{a'}\}$ \st $|A_3^t| = n-m$.\;
% 	 $A_1^t \defeq $ Set of $k$ arms with the highest empirical means.\;
%      $A_3^t \defeq $ Set of $n-m$ arms with the lowest empirical means.\;
%      $A_2^t \defeq \{\mathcal{A} \setminus (A_1^t \cup A_3^t)\}$.\;
%      $h_*^t = \arg \max_{\{a \in A_1^t\}} lcb(a,t)$.\;
%      $m_*^t = \arg \min_{\{a \in A_2^t\}} u_a^{t}$.\;
%      $l_*^t = \arg \max_{\{a \in A_3^t\}} ucb(a,t)$.\;
%      pull  $h_*^t,  m_*^t, l_*^t$.
%  }
%  \Return $A_1^t$.\;
%  }
% \end{algorithm}

\begin{algorithm}[ht]
\begin{algorithmic}
\small{
 \REQUIRE {$\mathcal{A}$ (\st $|\mathcal{A}| = n$), $k, m, \epsilon, \delta$.}
 \ENSURE {$k$ distinct $(\epsilon,m)$-optimal arms from $\mathcal{A}$.}
 \STATE Pull each arm $a  \in \mathcal{A}$ once. Set $t = n$.
 \WHILE {$ ucb({l_*^t}, t+1) - lcb({h_*^t}, t+1) > \epsilon.$} { \label{ln:stpkoutofm}
     \STATE $t = t + 1$.
     \STATE $A_1^t \defeq $ Set of $k$ arms with the highest empirical means.
     \STATE $A_3^t \defeq $ Set of $n-m$ arms with the lowest empirical means.
     \STATE $A_2^t \defeq \mathcal{A} \setminus (A_1^t \cup A_3^t)$.
     \STATE $h_*^t = \arg \max_{\{a \in A_1^t\}} lcb(a,t)$.
     \STATE $m_*^t = \arg \min_{\{a \in A_2^t\}} u_a^{t}$.
     \STATE $l_*^t = \arg \max_{\{a \in A_3^t\}} ucb(a,t)$.
     \STATE pull  $h_*^t,  m_*^t, l_*^t$.
 }\ENDWHILE
 \STATE Return $A_1^t$.
 }
\end{algorithmic}
\caption{\GLUCB: Algorithm to select $k$ $(\epsilon, m)$-optimal arms}
\label{alg:glucb}
\end{algorithm}

At each round $t$, we partition $\A$ into three subsets. We keep the $k$ arms
with the highest empirical averages in $A_1^t$, the $n-m$ arms with the lowest empirical averages in $A_3^t$,
and the rest in $A_2^t$; ties are broken arbitrarily (uniformly at random in our experiments). At each round we choose
a \emph{contentious} arm from each of these three sets: from 
$A_1^t$ we choose $h_*^t$,
the arm with the lowest lower confidence bound (LCB); from $A_2^t$ the arm which is least pulled is chosen, and called $m_*^t$; from $A_3^t$ we choose $l_*^t$, the arm with the highest
upper confidence bound (UCB). The algorithm stops as soon as the difference between the lower 
confidence bound of $h_*^t$, and the upper confidence bound of $l_*^t$ becomes no larger than 
the tolerance $\epsilon$.

Let $B_1, B_2, B_3$ be corresponding sets based on the true means: that is, subsets of $\mathcal{A}$ such that $B_1 \defeq \{1, 2,\cdots, k\}$,
$B_2 = \{k+1, k+2,\cdots, m\}$ and $B_3=\{m+1, m+2,\cdots, n\}$. For any two arms $a, b \in \mathcal{A}$ we define
$\Delta_{ab} \defeq \mu_a - \mu_b$. For the sake of convenience we slightly overload this notation as
{\footnotesize
\begin{equation}\label{eq:defdelta}
 \Delta_a = \begin{cases}
  \mu_a - \mu_{m+1}\; \text{if}\; a \in B_1\\
  \mu_k - \mu_{m+1}\; \text{if}\; a \in B_2\\
  \mu_m - \mu_a\;\;\;\;\; \text{if}\; a \in B_3.
 \end{cases}
\end{equation}
}
% \begin{table}[H]
% \centering
% \caption{My caption}
% \label{my-label}
% \begin{tabular}{l|lll}
% \hline
%  & $a \in B_1$ & $a \in B_2$ & $a \in B_3$ \\ \hline
% $\Delta_a$ & $\mu_a - \mu_{m+1}$ & $\mu_k - \mu_{m+1}$ & $\mu_m - \mu_a$ \\ \hline
% \end{tabular}
% \end{table}
We note that $\Delta_k = \Delta_{k+1} = \cdots = \Delta_m = \Delta_{m+1}$.
Let $u^*(a,t) \defeq \bceil{\frac{32}{\max\{\Delta_a, \frac{\epsilon}{2}\}^2}\ln\frac{k_1 n t^4}{\delta}}$ for all $a \in \mathcal{A}$, where $k_1=5/4$. 
Now, we define the hardness term as $H_\epsilon = \sum_{a \in \mathcal{A}}\frac{1}{\max\{\Delta_a, \epsilon/2\}^2}$.

\begin{restatable}{theorem}{thmscglucb}[Expected Sample Complexity of \GLUCB]
\label{thm:scglucb}
\GLUCB solves \QFK using an expected sample complexity upper bounded by
$O\left(H_\epsilon \log\frac{H_\epsilon}{\delta}\right)$. 
\end{restatable}
Appendix-A describes the proof in detail. The core argument 
is similar to
that for Algorithm $\F_2$ by \citet{bib:arcsk2017}. However, it subtly differs due to the different strategy for choosing arms since the output set is
not necessarily singleton.
%  Recently, \citet{Jamieson+N:2014} has shown that using a 
% tailored upper bound, \LUCB can be shown to incur an expected sample complexity
% which is within a $O(\log n)$ factor of the lower bound. A similar technique can
% also be adopted here to make a tighter analysis. However, in the interest of
% keeping the proof simple, we keep our analysis restricted in the conventional approach and leave the tighter analysis as a future exercise. 
In practice, one can use
tighter confidence bound calculations (we use KL-divergence based
confidence bounds in our experiments) to get even better sample complexity.
Next, we are going to consider infinite-armed bandit instances, and present the algorithms to solve them.
\section{Algorithm for Infinite Instances}
\label{sec:infinitemab}

Before proceeding to the identification of $k$ 
$[\epsilon, \rho]$-optimal arms in infinite-armed bandits, we revisit the case of $k = 1$. To find a single $[\epsilon, \rho]$-optimal arm,
the sample complexity of all the existing algorithms~\citep{bib:arcsk2017,Aziz+AKA:2018} scales as
$(1/\rho \epsilon^{2})\log^2(1/\delta)$, for the given mistake probability $\delta$. In this section
we present an algorithm \PPP whose sample complexity is only an \textit{additive} poly-log factor
away from the lower bound of $\Omega((1/\rho \epsilon^{2})\log 1/\delta)$~\citep[Corollary 3.4]{bib:arcsk2017}.
 
\subsection{Solving \protect\QP Instances}
\label{subsec:tighterqp}
\PPP is a two-phase algorithm.
In the first phase, it runs a sufficiently large number of independent copies of \PP and chooses
a large subset of arms  (say of size $u$), in which every arm is $[\epsilon, \rho]$-optimal 
with probability at least $1-\delta'$, where $\delta'$ is some small \textit{constant}. 
The value $u$ is chosen in a manner such that at least one of the chosen arms is 
$[\epsilon/2, \rho]$-optimal with probability at least $\delta/2$.
The second phase solves the best arm identification problem $(1,1,u)$ by applying \textsc{Median Elimination}.

Algorithm~\ref{alg:tightqp1} describes \PPP. It uses \PP~\citep{bib:arcsk2017} with  \textsc{ Median Elimination}  as a 
subroutine, to select an  $[\epsilon, \rho]$-optimal arm with confidence $1-\delta'$.
We have assumed $\delta' = 1/4$, in practice the one can choose any
sufficiently small value for it, which will merely affect the multiplicative constant in the upper bound.
% \begin{algorithm}[]
% \DontPrintSemicolon
% \small{
% \caption{$\mathcal{P}_3$}
%  \KwIn{ $\mathcal{A}, \epsilon, \delta$.}
% \label{alg:tightqp1}
%  Set $\delta' = 1/4$, $u = \bceil{\frac{1}{1-\delta'}\log\frac{2}{\delta}} = \bceil{\frac{4}{3}\log\frac{2}{\delta}}$.\;
%  Run $u$ copies of $\mathcal{P}_2(\A, \rho, \epsilon/2, \delta')$ and form set $S$ with the output arms.\;
%  Identify the $(\epsilon/2,1)$-optimal arm in $S$ using \textsc{Median Elimination} with confidence at least $1-\delta/2$.
% %  Return the output from \GHALVING$\left(S, u, \floor{\frac{u}{2}}, 1, \frac{\epsilon}{2}, \frac{\delta}{2}\right)$.
%  }
%  \end{algorithm}

\begin{algorithm}[ht]
\begin{algorithmic}
\small{
 \REQUIRE { $\mathcal{A}, \epsilon, \delta$.}
 \ENSURE {One $[\epsilon, \rho]$-optimal arm.}
 \STATE Set $\delta' = 1/4$, $u = \bceil{\frac{1}{\delta'}\log\frac{2}{\delta}} = \bceil{4\log\frac{2}{\delta}}$.
 \STATE Run $u$ copies of $\mathcal{P}_2(\A, \rho, \epsilon/2, \delta')$ and form set $S$ with the output arms.
 \STATE Identify an $(\epsilon/2,1)$-optimal arm in $S$ using \textsc{Median Elimination} with confidence at least $1-\delta/2$.
 }
 \end{algorithmic}
 \caption{$\mathcal{P}_3$}
\label{alg:tightqp1}
 \end{algorithm}

 \begin{restatable}{theorem}{thmppp}[Correctness and Sample Complexity of \PPP]
 \label{thm:ppp}
 \PPP  solves \QP, with sample complexity 
%  $O\left(\frac{1}{\epsilon^2}\left(\frac{1}{\rho}\log\frac{1}{\delta} + \log^2\frac{1}{\delta}\right)\right)$. 
$O(\epsilon^{-2}(\rho^{-1}\log(1/\delta) + \log^2(1/\delta)))$.
 \end{restatable}
 \begin{proof}
 First we prove the correctness and then upper-bound the sample complexity.

\paragraph{Correctness.} First we notice that each copy of $\mathcal{P}_2$ outputs an $[\epsilon/2, \rho]$-optimal arm
 with probability at least $1-\delta'$. 
%  Also, \GHALVING outputs an
%  $[\epsilon/2, \rho]$-optimal arm with probability $1-\delta$.
 Now, $S \cap \TOPRHO = \emptyset$ can only happen if all the $u$ copies of \PP output sub-optimal arms. Therefore, $\Pr\{S \cap \TOPRHO = \emptyset\} = (1-\delta')^{u} \leq \delta/2$.
%  Let, $\hat{X}$ be the fraction of sub-optimal arms in $S$. Then $\Pr\{\hat{X} \geq \frac{1}{2}\}$ $= \Pr\{\hat{X} - \delta' \geq \frac{1}{4}\}$
%   $\leq \exp(2\cdot(\frac{1}{4})^2\cdot u) = \exp(-2\cdot\frac{1}{16}\cdot 8\log\frac{2}{\delta}) < \frac{\delta}{2}$. 
  On the other hand, the mistake probability of \textsc{Median Elimination} is upper bounded by $\delta/2$. Therefore, by taking union bound, we get the 
  mistake probability is upper bounded by $\delta$. Also, the mean of the output arm is not
  less than $\frac{\epsilon}{2} + \frac{\epsilon}{2} = \epsilon$ from the $(1-\rho)$-th
  quantile.
  
  \paragraph{Sample complexity.} First we note that, for some appropriate constant $C$,
  the sample complexity (SC) of each of the $u$ copies of $\mathcal{P}_2$ is $\frac{C}{\rho(\epsilon/2)^2}\left(\ln\frac{2}{\delta'}\right)^2 \in O\left(\frac{1}{\rho\epsilon^2}\right)$.
  Hence, SC of all the $u$ copies $\mathcal{P}_2$ together is upper bounded by $\frac{C_1\cdot u}{\rho\epsilon^2}$, for some constant $C_1$.
  Also, for some constant $C_2$, the sample complexity of \textsc{Median Elimination} is upper bounded by $\frac{C_2\cdot u}{ (\epsilon/2)^2}\ln\frac{2}{\delta} \leq \frac{C_3}{\epsilon^2}\ln^2\frac{2}{\delta}$.
  Adding the sample complexities and substituting for $u$ yields the bound.
 \end{proof}
 
% Although, sample complexity of \PPP matches the lower bound up to a constant factor, in practice its performance is
% not satisfactory due to large value of the constant. 
% For example, the number of samples required to select $u$ $[\epsilon, \rho]$-optimal arms is 
% $\frac{C \cdot u}{\rho (\epsilon/2)^2} \log^2\frac{2}{\delta'} \geq \bceil{\frac{C}{\rho \epsilon^2}\log\frac{2}{\delta} \cdot 4\frac{\log^2(1/\delta')}{0.5-\delta'}} \geq \bceil{\frac{109 C}{\rho \epsilon^2}\log\frac{2}{\delta}}$, minimising with respect to $\delta' $.
% In the second phase a big number of samples add up to this to make it even bigger. 
% Therefore, \PPP can not outperform \PP if $\delta \geq 2^{-109}$, which is effectively  far below
% the practical range.

\begin{corollary}
\label{cor:qffromqptighter}
\PPP can solve any instance of \QF $(\A, n, m, \epsilon, \delta)$ with sample complexity
$O\left(\frac{1}{\epsilon^2}\left(\frac{n}{m}\log\frac{1}{\delta} + \log^2\frac{1}{\delta}\right)\right)$.
\end{corollary}
%\filler{NEED A COUPLE OF LINES DESCRIBING HOW.}
\begin{proof}
Let, $(\A, n, m, \epsilon,\delta)$ be the given instance of \QF.
We partition the set $\A^\infty = [0,1]$ in to $n$ equal segments and associate each 
with a unique arm in $\A$, and such that no two different subsets get associated
with the same arm. Now, defining $P_{A^\infty} = Uniform[0,1]$, and $\rho' = m/n$,
we realise that solving the \QP instance $(\A^\infty, P_{\A^\infty}, \rho', \epsilon,\delta)$
solves the original \QF instance, thereby proving the corollary.
 \end{proof}
% Given an instance of \QF by $(\A, n, m, \epsilon,\delta)$, we can transform it
% to an instance of \QP given by $(\A^\infty, P_{\A^\infty}, \rho', \epsilon,\delta)$, where
% $P_\A$ is an uniform distribution over $\A$, and $\rho = m/n$. Now, solving this \QP
% solves the original \QF, thereby proving Corollary~\ref{cor:qffromqptighter}.

At this point it is of natural interest to find an efficient algorithm to solve \QPK.
Next, we discuss the extension of \QP to \QPK, and present lower and  upper bound 
on the sample complexity needed to solve it.
% this in detail.
% However, as discussed in Section~\ref{sec:problemdefinitionandcontributions},
% problem instances have to be 

\subsection{Solving ``At Most $k$-equiprobable'' \protect\QPK Instances}
\label{subsec:tighterqpk}

Now, let us focus on identifying $k$ $[\epsilon, \rho]$-optimal arms.
In Theorem~\ref{thm:impossibility_qpk} we derive the lower bound on the sample complexity to solve an 
instance \QPK by reducing it to solving a \SUBSET problem as follows.

\begin{restatable}{theorem}{thmimpossibility_qpk}[Lower Bound on the Sample Complexity for Solving \protect\QPK]
\label{thm:impossibility_qpk}
% For $\alpha \in [0, 1)$, and $k > 1$,
For every $\epsilon \in (0, \frac{1}{\sqrt{32}}]$,
$\delta \in (0, \frac{1}{\sqrt{32}}]$,
and $\rho \in (0,\frac{1}{2}]$,
there exists an instance of \QPK given by $(\A, P_\A, \rho, \epsilon, \delta)$,
such that
any algorithm that solves \QPK incurs at least
%there exists no algorithm that solves it %that can solve every instance of \QPK given by $(\A, P_\A, \rho, \epsilon, \delta)$
%within a number of samples which is strictly lesser than 
$C\cdot \frac{k}{\rho\epsilon^2}\ln\frac{k}{8\delta}$ samples, where 
$C = \frac{1}{18375}$.
\end{restatable}
%\filler{THEOREM NEEDS TO BE WRITTEN BY SPECIFYING RANGES FOR EPSILON, DELTA, AND RHO.}
\begin{proof}
We shall prove the theorem by contradiction. Let us assume that  the
statement is incorrect. Therefore, there exists an algorithm \textsc{ALG} that \textsc{ALG} can solve
any instance of \QPK using no more than
$C\cdot \frac{k}{\rho\epsilon^2}\ln\frac{k}{8\delta}$ samples, for
$C= \frac{1}{18375}$. Now, let $(n, \A, m, \epsilon, \delta)$ be an instance of \SUBSET~\citep{bib:arcsk2017} with
$n \geq 2m$. Letting $P_\A = Uniform\{1,2, \dots, n\}$, $k = m$, and $\rho = m /n$, we
create an instance of \QPK as $(\A, P_\A, \rho, k, \epsilon, \delta)$. Therefore, solving
this \QPK instance will solve the original \SUBSET instance.
%Assuming $N$ as the number of samples needed by \textsc{ALG} to solve this \QPK instance,  we can write $N \leq C\cdot \frac{k}{\rho\epsilon^2}\ln\frac{k}{4\delta}$. Hence, a
According our claim, \textsc{ALG} solves the original \SUBSET instance using at most
$C\cdot \frac{k}{(k/n)\epsilon^2}\ln\frac{k}{8\delta}$
$ = C\cdot \frac{m}{(m/n)\epsilon^2}\ln\frac{m}{8\delta}$
$ = C\cdot \frac{n}{\epsilon^2}\ln\frac{m}{8\delta}$ samples. 
% Now,  for all $k > 1$, according to our assumption as $\alpha < 1$,
This observation contradicts the lower bound on the sample complexity for solving \SUBSET~\citep[Theorem 8]{bib:lucb}; thereby proving the theorem.
\end{proof}

% At this point it is of natural interest to find an efficient algorithm to solve \QPK. However, as discussed in Section~\ref{sec:problemdefinitionandcontributions}, problem instances have to be valid---that is, containing $k$ $[\epsilon, \rho]$-optimal arms. Even so, it can take an arbitrarily large number of guesses to discover unseen arms unless $P_\A$ allocates sufficient probability to each of $k$ arms. For this purpose, recall that we decided to limit our algorithms to at most $k$-equiprobable instances
% $(\A, P_\A, k, \rho, \epsilon, \delta)$ for which $\forall a \in \TOPRHO$, $\Pr_{\mathbf{a}' \sim P_\A}\{\mathbf{a}' = a\} \leq \frac{\rho}{k}$.

%\begin{definition}[At most equiprobable instance of \QPK:] Given a valid instance of \QPK as $(\A, P_\A, k, \rho, \epsilon, \delta)$, we call it \emph{at most equiprobable}, if $\forall a \in \TOPRHO$, $\Pr_{\mathbf{a}' \sim P_\A}\{\mathbf{a}' = a\} \leq \frac{\rho}{k}$.
%\end{definition}

\paragraph{Algorithm for solving at most $k$-equiprobable \protect\QPK instances.} Let, for any $\mathcal{S} \subseteq \A$, $\nu(\mathcal{S}) \defeq \Pr_{a \sim P_\A}\{a \in \mathcal{S}\}$. Therefore, $\nu(\A) = 1$.
Now, we present an algorithm \KQP that can solve any at most $k$-equiprobable instance of \QPK. 
Algorithm~\ref{alg:looseqpk} describes \KQP.
At each phase $y$, it solves an instance of \QP to output an arm, say $a^{(y)}$, from $\TOPRHO(\epsilon)$. In
the next phase, it updates the bandit instance $\A^{y+1} = \A^{y}\setminus\{a^{(y)}\}$,
the sampling distribution 
$P_{\A^{y+1}} = \frac{1}{1-\nu\left(\A\setminus\A^{y+1}\right)} P_{\A^{y}}$, and the target quantile $\rho^{y+1} = \rho^y-\nu(a^{(y)})$. However, as we
are not given the explicit form of $P_\A$, we realise $P_{\A^{y+1}}$ by rejection-sampling---if
$a' \in \A\setminus\A^{y+1}$ is chosen by $P_{\A}$, we simply discard $a'$, and
continue to sample $P_\A$ one more time. Because $\nu(\{a^y\})$ is not known explicitly,
we rely on the fact that $\nu(\{a^y\}) \leq \rho/k$: it is for this reason we require the instance to be at most $ k$-equiprobable.
Therefore, in each phase $y \geq 1$, $\rho^y - \rho/k \leq \rho^{y+1} \leq \rho^y - \nu\{a^y\}$, and hence,
\KQP solves an instance of \QP given by 
$\left(\A^{y}, P_{\A^{y}}, \rho-{(y-1)\rho}/{k}, \epsilon, \delta\right)$.

% We notice, in each phase the algorithm \KQP 
% queries the oracle with the arm output by \PPP.
% % As \PPP can solve any instance of \QP, 
% Therefore, if the original \QPK instance is at most
% uniform, we can pass an oracle as an input to \KQP which will produce $\nu(\{a^{(y)}\}) = \rho/k$ 
% for any $y \in \{1, \cdots, k\}$.
% For such an instance this oracle will help us to establish the upper bound on the
% expected sample complexity.
% \begin{algorithm}
% \caption{\KQP: Algorithm to solve a valid \QPK}
% \label{alg:looseqpk}
% \DontPrintSemicolon
% \KwIn{$\A, P_\A, k, \rho, \epsilon, \delta$, and an oracle that gives $ \Pr_{a \sim P_\A}\{a \in \mathcal{S}$, for any $\mathcal{S} \subseteq \A$.}
% \KwOut{Set of $k$ distinct arms from $\TOPRHO(\epsilon)$.}
% $\A^1 = \A, \rho^1 = \rho$.\;
% \For{$y = 1,2,3,\cdots,k$}{
% 	Run \PPP to solve the \QP instance given by
%     $(\A^y, P_{\A^y}, \rho^y, \epsilon, \frac{\delta}{k})$, and let $a^{(y)}$ be the output.\;
%     $\A^{y+1} = \A^{y}\setminus \{a^{(y)}\}$\;
%     $P_{\A^{y+1}} = \frac{1}{1-\nu\left(\A\setminus\A^{y+1}\right)} P_{\A^{y}}$.\;
%     $\rho^{y+1} = \rho^y-\nu(\{a^{(y)}\})$.\;
% }
% \end{algorithm}

\begin{algorithm}
\begin{algorithmic}
\small{
\REQUIRE {$\A, P_\A, k, \rho, \epsilon, \delta$.}
\ENSURE {Set of $k$ distinct arms from $\TOPRHO(\epsilon)$.}
\STATE $\A^1 = \A, \rho^1 = \rho$.
\FOR {$y = 1,2,3,\cdots,k$}{
    \STATE Run \PPP to solve the \QP instance given by
    \STATE $(\A^y, P_{\A^y}, \rho^y, \epsilon, \frac{\delta}{k})$, and let $a^{(y)}$ be the output.
    \STATE $\A^{y+1} = \A^{y}\setminus \{a^{(y)}\}$.
%     \STATE $P_{\A^{y+1}} = \frac{1}{1-\nu\left(\A\setminus\A^{y+1}\right)} P_{\A^{y}}$.
    \STATE $\rho^{y+1} = \rho^y-({(y-1)\rho})/{k}$.
}\ENDFOR
}
\end{algorithmic}
\caption{\KQP: Algorithm to solve a at most k-equiprobable \QPK instances}
\label{alg:looseqpk}
\end{algorithm}

In Theorem~\ref{thm:exsceqp} we present an upper bound on the expected sample complexity of \KQP.% \filler{NOT ADDRESSED: In the line just above the algorithm, and also in the algorithm, you are (incorrectly) subtracting $\rho/k$ from the existing $\rho^{y}$. I think you should instead subtract $\nu(\A^{(y)})$.} \textcolor{blue}{not clear to me what you mean}
\begin{theorem}
\label{thm:exsceqp}
Given any at most $k$-equiprobable instance of \QPK with $k > 1$, 
%  and an oracle that produces $\nu(\{a^{(y)}\}) = \rho/k$
%  for all $y \in  \{1, 2, \cdots, k\}$, \
 \KQP solves the instance with expected 
sample-complexity upper bounded by $O\left(\frac{k}{\epsilon^2}\left(\frac{\log k}{\rho}\log\frac{k}{\delta} + \log^2\frac{k}{\delta}\right)\right)$.
\end{theorem}
\begin{proof}
We break the proof in two parts: upper-bounding the sample complexity, and proving correctness.

\textbf{Sample complexity:} In phase $y$, the sample complexity of \PPP is upper-bounded as
$\text{SC}(y) \leq \frac{C}{\rho^y \epsilon^2}\log\frac{k}{\delta}$, for some constant $C$.
Therefore, the sample complexity of \KQP is upper bounded as 
{\small
\begin{align*}
& \sum_{y=1}^k \text{SC}(y) \leq \sum_{y=1}^k \frac{C}{ \epsilon^2}\left(\frac{1}{\rho^y}\log\frac{k}{\delta} + \log^2\frac{k}{\delta}\right),\\
%  &= \left(\frac{1}{\rho} + \sum_{y=2}^k \frac{1}{\rho - \sum_{j=1}^{y-1} \nu(\{a^{(j)}\})}\right) \frac{C}{\epsilon^2}\ln\frac{k}{\delta} +\\
%  &\hspace{1cm}  \frac{kC}{\epsilon^2}\ln^2\frac{k}{\delta},\\
& \leq  \frac{C}{\epsilon^2}\left(\log\frac{k}{\delta} \sum_{y=1}^k \frac{1}{\rho - (y-1)\frac{\rho}{k}}+ k\log^2\frac{k}{\delta}\right),\\
& = \frac{Ck}{\epsilon^2} \left(\frac{1}{\rho} \log\frac{k}{\delta} \sum_{y=1}^k \frac{1}{k-y+1} + \log^2\frac{k}{\delta}\right),\\
& \leq \frac{C'k}{\epsilon^2}\left(\frac{\log k}{\rho}\log\frac{k}{\delta} + \log^2\frac{k}{\delta}\right),
\end{align*}
}
for $k > 1$, and some constant $C'$.
\paragraph{Correctness:} 
Letting $E_y$ be the event that $a^{(y)} \not\in \TOPRHO(\epsilon)$, the probability of mistake by \KQP can be upper bounded as $\Pr\{\text{Error}\} = \Pr\{\exists y \in \{1, \cdots, k\}\; E_y \} \leq \sum_{y=1}^k \Pr\{E_y \} \leq \sum_{y=1}^k \frac{\delta}{k} = \delta$.
\end{proof}

\begin{corollary}
\label{cor:qfkfromqpktighter}
\KQP can solve any instance of \QFK given by $(\A, n, m, k, \epsilon, \delta)$ with $k \geq 2$, using $O\left(\frac{k}{\epsilon^2}\left(\frac{n\log k}{m}\log\frac{k}{\delta} + \log^2\frac{k}{\delta}\right)\right)$ samples.
\end{corollary}

We note that though the sample complexity of \KQP is independent of size of the bandit instance $\A$, and every
instance of \QFK given by $(\A, n, m, m , \epsilon, \delta)$, can be solved by \KQP by posing it as an instance of 
\QPK given by $(\A,  Uniform\{\A\}, m/n, m, \epsilon, \delta)$. However, for $k=m$, the sample complexity of \KQP
reduces to $O\left(\frac{1}{\epsilon^2}\left(n\log m\cdot\log\frac{m}{\delta} + \log^2\frac{m}{\delta}\right)\right)$,
which is higher than the sample complexity of \HALVING~\cite{bib:explorem}, that needs only  $O\left(\frac{n}{\epsilon^2}\log\frac{m}{\delta}\right)$ samples.
Hence, for the best subset selection problem 
in finite instances \HALVING is preferable to \KQP.
However, in the very large instances, where the probability
of picking any given arm from $\TOPRHO$ is close to zero,
% However,
% for potentially infinite instances, where the probability
% of picking any given arm is zero (or very close to zero),
\QPK is the ideal problem to solve, and \KQP is the first
solution that we propose.

\begin{corollary}
\label{cor:qfkfromqpktighter2}
Every instance of \QPK given by $(\A, P_\A, k, \rho, \epsilon, \delta)$, such that
$|\A| = \infty$, %and $\Pr_{a\sim P_\A}\{a \in S \subset \A : |S| < \infty\} = 0$,
and for all finite subset $S \subset \A$,  $\Pr_{a\sim P_\A}\{a \in S\} = 0$;
can be solved within a sample-complexity $O\left({k}{\epsilon^{-2}}\left({\rho^{-1}}\log({k}/{\delta}) + \log^2({k}/{\delta})\right)\right)$, by independently solving $k$ different \QP instances, each given by $(\A, P_\A, k, \rho, \epsilon, \delta/k)$.
\end{corollary}
The correctness of Corollary~\ref{cor:qfkfromqpktighter2} gets proved by noticing the fact that all the $k$ outputs are unique with probability 1, and then taking union bound over mistake probabilities.
% As the probability of encountering any arm more than once is zero, the each of the $k$ outputs is
% is unique with probability 1. Now, taking union over mistake probabilities, the statement of Corollary~\ref{cor:qfkfromqpktighter2} follows.
%
%\filler{Still not clear. Are you assuming $k = m$? Perhaps you need to specify the constraints on $k$ and $m$. Otherwise, look at the preceding sentence alone, which implies $(1, m, n)$ is better solved using HALVING unless $n = \infty$. Not true; what if $n$ is just large?}
Before going to the experiments, we present an important result on the hardness of solving \QP.
% Specifically,
% we establish a connection between the optimal sample complexities for solving \QF and \QP.
% \filler{Tight lower bound or tight upper bound?}

\subsection{On the Hardness of Solving \protect\QP}
\label{subsec:qpreducub}
% \filler{What do you mean by ``order-optimal''? The term suggests that you believe that the lower bound we have furnished is correct. But what if the actual lower bound is larger?}

Theorem~\ref{thm:qpreducubonly} presents a general relation between the upper bound on 
sample complexities for solving \QF and \QP.
 
% solving \QP problem using order-optimal sample complexity is crucially dependent on the
% existence of an algorithm for solving \QF within order-optimal sample complexity.
% The lower bound directly follows from the intuition. We provide the upper bound in Theorem~\ref{thm:qpreducub}.
% \filler{Not clear: lower bound for what and upper bound for what?}

% if every problem instance of
% \QP given by $(\A, P_\A, \rho, \epsilon, \delta)$   needs $\Theta(h(\rho,\epsilon,\delta))$ 
% samples to get solved.
%
% \subsection{Upper Bound}
% \label{subsec:hardnessqpub}

% In this section we show that if there exists an algorithm for solving any instance of \QF
% within order-optimal sample-complexity, then we can construct an algorithm that can solve
% any instance of \QP  within order-optimal sample-complexity.

\begin{restatable}{theorem}{thmqpreducubonly}
\label{thm:qpreducubonly}
Let $\gamma: \mathbb{Z}^+ \times \mathbb{Z}^+ \times [0,1] \times [0,1] \mapsto \mathbb{R}^+$.
If every instance of \QF  given by $(\A, n, m, \epsilon, \delta)$, can be solved
within the sample-complexity $O\left(\frac{n}{m\epsilon^2}\log\frac{1}{\delta} + \gamma(n,m,\epsilon,\delta)\right)$, 
then,
% there exists an algorithm \textsc{OptQP} that can solve 
every instance of \QP  given by $(\A, P_\A, \rho, \epsilon, \delta)$ can be solved
within the sample-complexity  
\resizebox{\columnwidth}{!}{$O\left({(1/\rho\epsilon^{2}})\log({1}/{\delta}) + \gamma\left(\ceil{8\log({2}/{\delta})}, \floor{4\log({2}/{\delta)}}, {\epsilon}/{2}, {\delta}/{2}\right)\right)$.}
\end{restatable}

We assume that there exists an algorithm \textsc{OptQF} that solves
every instance of \QF  given by $(\A, n, m, \epsilon, \delta)$,
using $O\left(\frac{n}{m\epsilon^2}\log\frac{1}{\delta} + \gamma(n,m,\epsilon,\delta)\right)$ samples.
% We prove Theorem~\ref{thm:qpreducub}
We establish the upper bound on sample complexity for solving \QP by constructing an  algorithm \textsc{OptQP}
that follows an approach similar to \PPP. Specifically, \textsc{OptQP} reduces the input \QP instance 
to an instance of \QF 
using $O\left(\frac{1}{\rho\epsilon^2}\log\frac{1}{\delta}\right)$ samples. Then, it solves that 
\QF  using \textsc{OptQF} as the subroutine. The detailed proof is given in Appendix-C.

\begin{corollary}
\label{cor:effectoptqf}
Corollary~\ref{cor:qffromqptighter} shows that every \QF is solvable in $O\left(\frac{1}{\epsilon^2}\left(\frac{n}{m}\log\frac{1}{\delta} + \log^2\frac{1}{\delta}\right)\right)$ samples.
Hence, $\gamma(n,m,\epsilon,\delta) \in O\left(\frac{1}{\epsilon^2}\log^2\frac{1}{\delta}\right)$,
and therefore, every \QP is solvable in $O\left(\frac{1}{\epsilon^2}\left(\frac{1}{\rho}\log\frac{1}{\delta} + \log^2\frac{1}{\delta}\right)\right)$ samples.

On the other hand, if the lower bound for solving \QF provided by \citet{bib:arcsk2017} matches the upper bound up to a constant factor, then $\gamma(n,m,\epsilon,\delta) \in \Theta\left(\frac{n}{m\epsilon^2}\log\frac{1}{\delta}\right)$. In that case, \QP is solvable using $\Theta\left(\frac{1}{\rho\epsilon^2}\log\frac{1}{\delta}\right)$ samples.
\end{corollary}

It is interesting to find a $\gamma(\cdot)$ such that the upper bound presented in Theorem~\ref{thm:qpreducubonly}
matches the lower bound up to a constant factor. We notice, Theorem~\ref{thm:qpreducubonly} guarantees that there exists a constant $C$,
such that for any given $\epsilon, \delta$, and $m \leq n/2$,
%for every $n \geq \ceil{8\log(2/\delta)}$, %and for every $m \leq n/2$,
$\gamma(n ,m, \epsilon, \delta) \leq C \cdot \gamma\left(\ceil{8\log(2/\delta)}, \floor{4\log(2/\delta)}, \frac{\epsilon}{2}, \frac{\delta}{2}\right)$. However, for $n < \ceil{8\log(2/\delta)}$ %and $m \leq n/2$,
we believe \QF can be solved more efficiently than posing it as \QP. We present it as a conjecture.

\begin{definition}
For $g: \mathbb{Z}^+ \times \mathbb{Z}^+ \times [0,1] \times [0,1] \mapsto \mathbb{R}^+$ we say \QF is solvable in $\Theta(g(\cdot))$, if there exists an algorithm that
solves every instance of \QF given by $(\A, n, m, \epsilon, \delta)$ in $O(g(n,m,\epsilon,\delta))$ samples,
and there exists an instance of \QF given by $(\bar{\A}, \bar{n}, \bar{m}, \bar{\epsilon}, \bar{\delta})$ such that every algorithm
needs at least $\Omega(g(\bar{n},\bar{m},\bar{\epsilon},\bar{\delta}))$ samples to solve it.
\end{definition}

\begin{restatable}{conjecture}{conjdiffqfqp}
There exists a constant $C > 0$, and functions $g: \mathbb{Z}^+ \times \mathbb{Z}^+ \times [0,1] \times [0,1] \mapsto \mathbb{R}^+$, and $h: \mathbb{Z}^+ \times \mathbb{Z}^+ \times [0,1] \times [0,1] \mapsto \mathbb{R}^+$,
such that for every $\delta \in (0, 1]$, there exists an integer $n_0 <  C\log\frac{2}{\delta}$, such that for
every $n\leq n_0$, \QF 
%given by $(\A, n, m, \epsilon, \delta)$ 
is solvable in $\Theta(g(n, m, \epsilon, \delta))$ samples, and its  equivalent \QP (obtained by posing the
the instance of \QF as an instance of \QP, as done in proving Corollary~\ref{cor:qffromqptighter}) needs at least  $\Omega(h(n, m, \epsilon, \delta))$ samples, then $\lim_{\delta \downarrow 0} \frac{g(n, m, \epsilon, \delta)}{h(n, m, \epsilon, \delta)} \to 0$.
\end{restatable}

% Analogously, for a function $h: [0,1] \times [0,1] \times [0,1] \mapsto \mathbb{R}^+$ we say \QP is solvable 
% in $\Theta(h(\cdot))$, if there exists an algorithm \textsc{OptQP} that
% solves every instance of \QP given by $(\A, P_\A, \rho, \epsilon, \delta)$ in $O(h(\rho,\epsilon,\delta))$ samples, and there exists an instance of \QP given by $(\A', P_{\A'}', \rho', \epsilon', \delta')$ such that every algorithm needs at least $\Omega(h(\rho',\epsilon',\delta'))$ samples to solve it.

% \begin{corollary}
% \label{cor:effectoptqf}
% If the lower bound provided by \citet{bib:arcsk2017} is tight up to a constant
% factor, then $\gamma(n,m,\epsilon,\delta) \in \Omega\left(\frac{n}{m\epsilon^2}\log\frac{1}{\delta}\right)$. In that case, \QP is solvable using $\Theta\left(\frac{1}{\rho\epsilon^2}\log\frac{1}{\delta}\right)$ samples.
% On the other hand, if \QF is solvable in $\Theta\left(\frac{1}{\epsilon^2}\left(\frac{n}{m}\log\frac{1}{\delta} + \log^2\frac{1}{\delta}\right)\right)$ samples, then 
% $\gamma(n,m,\epsilon,\delta) \in \Omega\left(\frac{1}{\epsilon^2}\log^2\frac{1}{\delta}\right)$,
% and hence, solving \QP is solvable in $\Theta\left(\frac{1}{\epsilon^2}\left(\frac{1}{\rho}\log\frac{1}{\delta} + \log^2\frac{1}{\delta}\right)\right)$ samples.
% \end{corollary}

% \filler{``If the lower bound provided by \citet{bib:arcsk2017} is tight up to a constant
% factor, then $\gamma(n,m,\epsilon,\delta) = C\frac{n}{m\epsilon^2}\log\frac{1}{\delta}$,
% for some constant $C$.'' Do you mean $\gamma = \Omega(whatever)$ or $\geq C \cdot whatever$?}

Next, we empirically compare \GLUCB for 
$k=1$ with \FF on different instances, and also we study empirical performance of \GLUCB by varying $k$.

\section{Experiments and Results}
\label{sec:expt}
We begin our experimental evaluation by comparing  \FF~\cite{bib:arcsk2017} and \GLUCB based on the number
of samples drawn on different instances of \QF or $(1, m, n)$. \FF is a fully-sequential algorithm that resembles \GLUCB, but subtle differences in the way the algorithms partition $\A$ and select arms to pull lead to different results. At each time step $t$, \FF creates three partitions of $\A$---$\bar{A_1}(t)$, $\bar{A_{2}}(t)$, and $\bar{A_{3}}(t)$.
It puts the arm with the highest LCB in $\bar{A_1}(t)$; among
the rest, it puts $m-1$ arms with the highest UCBs in $\bar{A_2}(t)$; and the rest $n-m$ arms in $\bar{A_3}(t)$; ties are broken at random. At each time step $t$, it 
samples three arms---the arm in $\bar{A_1}(t)$, the least sampled arm in $\bar{A_2}(t)$, and the arm with the highest UCB in $\bar{A_3}(t)$.

% This lead \FF to sample the arms which are not the most 
% contentious for many times. Thus the most contentious arm needs 
% to wait longer to get the required number of pulls.  On the other hand, 
% the arm choosing rule in \GLUCB can closely guess the contentious arms.

\begin{figure}[h]
\centering
 \includegraphics[height=1.2in]{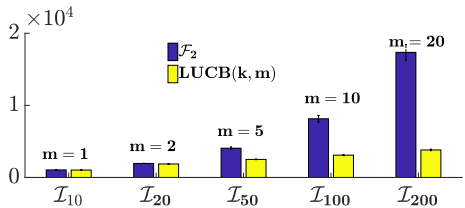}
 \caption{Comparison of incurred sample complexities by \FF and \GLUCB 
	  to solve \QF with  $m= 0.1 \times n$, on the five instances
	  detailed in Section~\ref{sec:expt}.
	  y-axis represents the number of samples averaged over 100 runs, with standard error bars.}
 \label{fig:scf2glucb}
\end{figure}

% \begin{table}[]
% \centering
% \caption{A comparison among the bandit instances.
% Columns from the left to right represent he number of arms,
% difference of means of consecutive arms, the number of
% $(\epsilon, m)$-optimal arms; and the hardness (defined in the Equation~\ref{eq:hardness}), $H_\epsilon$,
% for $\epsilon= 0.05$, and $k = 1$.}
% \label{tab:data_desc}
% \begin{tabular}{|l|r|r|r|}
% \hline
% \multicolumn{1}{|c|}{$n$} & \multicolumn{1}{c|}{$\mu_i - \mu_{i+1}$} & \multicolumn{1}{c|}{\begin{tabular}[c]{@{}c@{}}$\mathcal{TOP}_m(\epsilon)$\\ $m = 0.1\cdot n$\end{tabular}} & \multicolumn{1}{c|}{\begin{tabular}[c]{@{}c@{}}$H_\epsilon$\\ $m = 0.1\cdot n$\end{tabular}} \\ \hline
% 10 & 0.1109 & 1 & 206.556 \\ \hline
% 20 & 0.0525 & 2 & 396.396 \\ \hline
% 50 & 0.0204 & 7 & 966.957 \\ \hline
% 100 & 0.0101 & 14 & 1920.006 \\ \hline
% 200 & 0.0050 & 29 & 3827.218 \\ \hline
% \end{tabular}
% \end{table}
We take five Bernoulli instance of  sizes $n = 10, 20, 50, 100$, and $200$, with  the means linearly
spaced between 0.999 and 0.001  (both inclusive), and sorted in descending order.
% as the
% highest and the lowest means, and the rest are linearly spaced
% We take five bandit instances of sizes $n = 10, 20, 50, 100$, and $200$.
% In each of them, the highest and the lowest mean rewards are 0.999 and 0.001 respectively. Also, $\mu_{a_i} > \mu_{a_j}$ whenever $1 \leq i < j \leq n$, and for all $i \in \{1, \cdots, n-2 \}$, $\mu_i - \mu_{i+1} = \mu_{i+1} - \mu_{i+2}$.
We name the bandit instance of size $n$ as $\I_n$.
Now, setting $\epsilon = 0.05, \delta = 0.001$, and $m = 0.1 \times n$,
we run the experiments and compare the number of samples drawn by \FF and \GLUCB to solve
these five instances for $k=1$.
In our implementation we have used KL-divergence based confidence bounds~\cite{bib:klucb1,bib:klucb2} for both \FF and \GLUCB.
As depicted by Figure~\ref{fig:scf2glucb}, as the number of arms (n) increases,
the sample complexity of both the algorithms increases due to increase
in hardness $H_\epsilon$.
% (the fourth column of Table~\ref{tab:data_desc} in Appendix~\ref{app:expt}).
However, the sample complexity of \FF increases much faster than \GLUCB.

\begin{figure}[h]
\centering
 \includegraphics[height=1.5in]{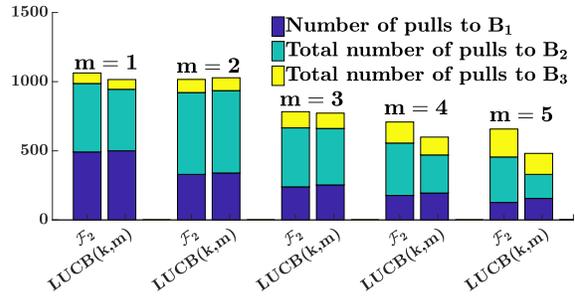}
 \caption{Comparison between \FF and \GLUCB on the number of
	  pulls received by the camps $B_1, B_2$ and $B_3$, for solving
	  different instances of \QF on $\I_{10}$, by varying $m$ from 1 to 5. 
	  Recall that $B_1$ is the singleton set, with the best arm being 
	  the only member. y-axis represents the number of samples averaged over 100 runs.}
 \label{fig:fracoptpull}
\end{figure}

As shown by \citet{Jamieson+N:2014} the efficiency of \LUCB comes from the quick identification of the most 
optimal arm due to a large separation from the $m+1$-th arm.
Intuitively, the possible reason for \FF to incur more samples is the delay in prioritising the 
optimal arm to pull more frequently. This should result in a smaller fraction of total samples taken from the best 
arm. Figure~\ref{fig:fracoptpull} affirms this intuition. It represents a comparison between \FF and \GLUCB on the number of samples obtained by
three ``ground-truth'' groups---$B_1$, $B_2$, and $B_3$
on $\I_{10}$, keeping $k=1$ and varying $m$ from 1 to 5. We note that
the lesser the difference between $k$ and $m$, the higher the hardness ($H_\epsilon$), and
both \FF and \GLUCB find it hard to identify a correct arm. Hence, for $k = m =1$, both of them
spend almost the same fraction of pulls to the best arm.
However, as $m$ becomes larger, keeping $k=1$, the hardness of the problem reduces, but
\FF still struggles to identify the best arm and results in spending a significantly
a lesser fraction of the total pulls to it, compared to \GLUCB. 

\begin{figure}[h]
\centering
 \includegraphics[height=1.2in]{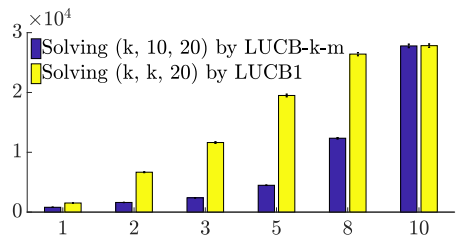}
 \caption{ Comparison of number of samples incurred for 
%  Comparison of sample complexities of incurred samples by \GLUCB for 
	  solving different  instances of \QFK defined on $\I_{20}$, by 
	  setting $m = 10$, and varying $k \in \{1, 2, 3, 5, 8, 10\}$.
	  x-axis represents $k$, and y-axis represents the number of samples 
	  averaged over 100 runs, with standard error bars.}
 \label{fig:vark_fixtopm}
\end{figure}

In this paper, we have developed algorithms specifically for the \QFK problem; previously one might have solved \QFK either by solving $(k, k, n)$ or $(m, m, n)$: that is choosing the \textit{best} $k$- or $m$-sized subset.  In Figure~\ref{fig:vark_fixtopm} we present a comparison of the sample complexities 
for solving  \QFK and the best subset-selection problems. 
Fixing $\A = \I_{20}$, $n=20$, $m = 10$, \QFK instances 
are given by and varying $k \in \{1, 3, 5, 8, 10\}$, whereas, for the best subset-selection problem we set $m =k$.
As expected, the number of samples incurred is significantly lesser for solving the problem instances with $k < m$, thereby validating the use of \GLUCB. %\filler{Best to place figures and tables Top or Bottom rather than Here.}

% Recalling the fact that for $k = m$, \GLUCB exactly coincides with \LUCB,  in Figure~\ref{fig:vark_fixtopm}
% we present a comparison of their complexities
% on $\I_{20}$, fixing $m = 10$, and varying $k \in \{1, 3, 5, 8, 10\}$. As expected, the number of samples 
% incurred by \GLUCB for solving \QFK with $k < m$, is far below that for solving $k = m$.

% We study how does the number of samples by \GLUCB increases with $k$. Considering $\I_{20}$, fixing $m = 10$, 
% for $k \in \{1, 3, 5, 8, 10\}$,
% we plot the number of samples incurred by \GLUCB in Figure~\ref{fig:vark_fixtopm}.

% {\color{blue} By definition, \GLUCB exactly coincides with LUCB for $k == m$.).}

% We notice that for $k = 1, 2, 3, 5, 8$, and $10$, $H_\epsilon = 396.4, 1029.7, 1428.6,
% 2225.9, 3420.3$, and $4214.8$ respectively. Therefore, 
% as supported by the theory, the number of 
% samples drawn by \GLUCB increases rapidly as $k$ gets closer to $m$.

\section{Conclusion}
\label{sec:conclusion}
Identifying one arm out of the best $m$, in an $n$-armed stochastic bandit
is an interesting problem identified by \citet{bib:arcsk2017}. They have
mentioned the scenarios where identifying the best subset is practically infeasible.
However, there are numerous examples in practice that demand efficient identification
of multiple good solutions instead of only one; for example, assigning a distributed crowd-sourcing task, identification of good molecular combinations in drug designing, \etc. 
In this paper, we present \QFK---a generalised problem of identifying $k$ out of the best $m$ arms. Setting $k=1$, \QFK gets reduced to selection of one out of the best $m$ arms, while
setting $k=m$, makes it identical with the ``subset-selection''~\cite{bib:explorem}. We
have presented a lower bound on the sample complexity
to solve \QFK. 
% Besides, being motivated by \HALVING~\cite{bib:explorem},
% we have presented an algorithm \GHALVING that solves \QFK with worst case sample complexity
% within a constant factor of the lower bound. 
We have also presented a fully sequential adaptive PAC algorithm, \GLUCB,
that solves \QFK, with expected sample complexity matching up to a constant factor that of 
$\F_2$~\cite{bib:arcsk2017} and \LUCB~\cite{bib:lucb} for $k=1$ and $k=m$, respectively. We have empirically
compared \GLUCB to $\F_2$ on different problem instances, and
shown that \GLUCB outperforms $\F_2$ by a large margin in terms of the number of
samples as $n$ grows.

For the problem of identification of a single $[\epsilon, \rho]$-optimal~\cite{bib:arcsk2017}
arm in infinite bandit instances, the existing upper bound on the sample complexity
differs from the lower bound by a multiplicative $\log\frac{1}{\delta}$ factor. 
It was not clear whether the lower bound was loose, or the upper can be improved,
and left as an interesting problem to solve~\cite{Aziz+AKA:2018}. 
% In this paper, we bridge the gap by proposing a non-adaptive algorithm $\mathcal{P}_3$, thus eliminating the extra $\log\frac{1}{\delta}$ factor from the upper bound. 
In this paper we reduce the gap by furnishing an upper bound which is optimal up to an \textit{additive} poly-log term.
Further, we show that the problem of identification k distinct $[\epsilon, \rho]$-optimal 
arms is not well-posed in general, but when it is, we derive a lower bound on the sample complexity. 
Also, we identify a class of well-posed instances for which we present an efficient algorithm. 
In the end we show that how improving the upper bound on the sample-complexity for solving 
\QF instances can be translated in improving  upper bound on the 
sample-complexity for solving \QP. However, we conjecture that there exists a set of \QF instances 
and a corresponding set of \QP instances, such that every instance of \QF requires lesser number of samples to solve
than the corresponding \QP instance in the other set. Showing correctness of the conjecture and improving the
lower and the upper bound on the sample complexities are some interesting directions we leave for future work.
% In the end
% we show that existence of an algorithm that solves \QF within a order-optimal sample complexity ensures 
% the existence of algorithm that solves \QP within a order-optimal sample complexity. 
% Finding such an algorithm is a very interesting problem that we leave as a future work. 

%{\footnotesize
\bibliographystyle{icml2019} %plainnat
\bibliography{rpbibfile}
%}
\pagebreak
\begin{appendices}
\onecolumn
\section{ Lower Bound on the Worst Case Sample Complexity to Solve \QFK}
 \label{app:lowerboundqfk}
\thmlbmainthm*
%%%%%%%%%%%%%%%

The proof technique for Theorem~\ref{thm:lbmainthm} follows a path similar to
that of~\citep[Theorem 8]{bib:lucb}, but
differs in the fact that any $k$ of the $m$ $(\epsilon, m)$-optimal arms
 needs to be returned as opposed to all the $m$. 
%  For convenience, we break the proof into
%  two parts. First, we prove it for the problem instances where either $m \leq 2k -2$ or
%  $m \geq 2k$; then we show that if the lower bound holds for those cases, it must hold
%  for $m= 2k-1$ for a sufficiently small constant.

\subsection{Bandit Instances:}
Assume we are given a set of $n$ arms $\mathcal{A} = \{0, 1, 2, \cdots, n-1\}$.
Let $I_0 \defeq \{0, 1, 2, \cdots, m-k\}$ and
$\I_l \defeq \{I : I\subseteq \{\mathcal{A}\setminus I_0\} \wedge |I| = l\}$.
Also for $I \subseteq \{m-k+1, m-k+2,\cdots,n-1\}$, we define $$\bar{I} \defeq \{m-k+1, m-k+2,\cdots,n-1\} \setminus I.$$

With each $I \in \I_{k-1} \cup \I_m$ we associate an $n$-armed bandit instance $\mathcal{B}^I$,
in which each arm $a$ produces a reward from a Bernoulli distribution with mean
$\mu_a$ defined as:
  \begin{align}
   \mu_a = \begin{cases}
            \frac{1}{2} & \text{ if } a \in I_0\\
        \frac{1}{2} + 2\epsilon & \text{ if } a \in I\\
        \frac{1}{2} - 2\epsilon & \text{ if } a \in \bar{I}.
           \end{cases}
  \end{align}
 Notice that all the instances in $ \I_{k-1} \cup \I_m$ have exactly $m$ $(\epsilon, m)$-optimal
 arms. For $I \in \I_{k-1}$, all the arms in $I_0$ are $(\epsilon, m)$-optimal, but for 
 $I \in \I_m$ they are not. With slight overloading of notation we write 
$\mu(S)$ to denote the multi-set consisting of means of the arms in $S\subseteq \A$.

 The key idea of the proof is that without sufficient  sampling
 of each arm, it is not possible to correctly identify $k$ of the $(\epsilon, m)$-optimal
 arms with high probability.
 
\subsection{Bounding the Error Probability:}
% \textbf{Bounding Error Probability}:
We shall prove the theorem by first making the following assumption, which we
shall demonstrate leads to a contradiction.

\begin{assumption}\label{asmp:contra} 
Assume, that there exists an algorithm $\mathcal{L}$, that solves each problem instance %$(\A, n, m, k, \epsilon, \delta)$
in \QFK defined on bandit instance $\mathcal{B}^I,\; I \in \I_{k-1}$,
 and incurs a sample complexity $\textsc{SC}_I$. Then for all $I \in \I_{k-1}$, 
 $\Ex{\textsc{SC}_I} < \frac{1}{18375}. \frac{1}{\epsilon^2}. \frac{n}{m-k+1}\ln\left(\frac{\binom{m}{m-k+1}}{4\delta}\right)$,
 for $0< \epsilon \leq \frac{1}{\sqrt{32}}$, $0 < \delta \leq \frac{e^{-1}}{4}$, and $n \geq 2m$, where $C = \frac{1}{18375}$.
\end{assumption}

For convenience, we denote by $\Pr_I$ the probability distribution induced by
the bandit instance $\mathcal{B}^I$ and the possible randomisation introduced
by the algorithm $\mathcal{L}$. Also, let $S_{\mathcal{L}}$ be the set of arms returned (as output) by
$\mathcal{L}$, and $T_S$ be the total number of times the arms in  $S \subseteq \mathcal{A}$ get sampled
until $\mathcal{L}$ stops.

Then, as $\mathcal{L}$ solves \QFK, for all $I \in \I_{k-1}$
\begin{equation}
\label{eq:contra}
 \Pr_I\{S_\mathcal{L} \subseteq I_0 \cup I\} \geq 1 - \delta.
\end{equation}

Therefore, for all $I \in \I_{k-1}$
\begin{equation}
 \mathbb{E}_{I}[T_\mathcal{A}] \leq C \frac{n}{m-k+1} \ln\left(\frac{\binom{m}{m-k+1}}{4\delta}\right).
\end{equation}

\subsubsection{Changing $\Pr_{I}$ to $\Pr_{I \cup Q}$ where $Q \in \bar{I}$ s.t. $|Q| = m-k+1$: }
Consider an arbitrary but fixed $I \in \I_{k-1}$. Consider a fixed partitioning of $\A$, into $\bfloor{\frac{n}{m-k+1}}$
subsets of size $(m-k+1)$ each.
If Assumption~\eqref{asmp:contra} is correct, then 
for the instance $\mathcal{B}^{I}$,
there are at most $\bfloor{\frac{n}{4(m-k+1})}-1$ partitions $B \subset\bar{I}$, such that $\mathbb{E}_{I}\left[T_B\right] \geq \frac{4C}{\epsilon^2}\ln\left(\frac{1}{4\delta}\right)$.
Now, as $\bfloor{\frac{n-m}{m-k+1}} - \left(\bfloor{\frac{n}{4(m-k+1)}} - 1\right)$ 
% $ > \bfloor{\frac{n}{2(m-k+1)}} - \frac{n}{4(m-k+1)}$
$\geq \bfloor{\frac{n}{4(m-k+1)}} + 1 > 0$;
therefore, there exists at least one subset $Q \in \bar{I}$ such that
$|Q| = m-k+1$, and $\mathbb{E}_{I}\left[T_Q\right] < \frac{4C}{\epsilon^2}\ln\left(\frac{\binom{m}{m-k+1}}{4\delta}\right)$.
Define $T^* = \frac{16C}{\epsilon^2}\ln\left(\frac{\binom{m}{m-k+1}}{4\delta}\right)$. Then
using Markov's inequality we get:
\begin{equation}\label{eq:lbusage}
 \Pr_{I}\left\{T_Q \geq T^*\right\} < \frac{1}{4}.
\end{equation}

Let $\Delta = 2\epsilon T^* + \sqrt{T^*}$ and also let $K_Q$ be the total rewards obtained from $Q$.
\begin{lemma}\label{lem:boundreward}
If $I \in \I_{k-1}$ and $Q \in \bar{I}$ s.t. $|Q| = m-k+1$, then 
 $$\Pr_{I}\left\{T_{Q} \leq T^* \wedge K_{Q} \leq \frac{T_{Q}}{2} - \Delta\right\} \leq \frac{1}{4}\;.$$
\end{lemma}
\begin{proof}
 Let $K_{Q}(t)$ be the total sum obtained from $Q$ at the end of
 the trial $t$. As for $\mathcal{B}^{I_0},\; \forall j\in Q\; \mu_j = 1/2 -2\epsilon$,
 hence selecting and pulling one arm at each trial from $Q$ following any rule (deterministic or probabilistic) is
 equivalent to selection of a single arm from $Q$ for once and subsequently perform
 pulls on it. Hence whatever be the strategy of pulling one arm at each trial from $Q$, the
 expected reward for each pull will be $1/2-2\epsilon$. Let $r_i$ be the i.i.d. reward
 obtained from the $i^\text{th}$ trial. Then $K_{Q}(t) = \sum_{i=1}^t r_i$ and
 $Var\left[r_i\right] = \left(\frac{1}{2} - 2\epsilon\right) \left(\frac{1}{2} + 2\epsilon\right) = \left(\frac{1}{4} - 4\epsilon^2\right) < \frac{1}{4}$.
 As $\forall i: 1 \leq i \leq t$, $r_i$ are i.i.d., we get $Var[K_{Q}(t)] = \sum_{i=1}^tVar(r_i) < \frac{t}{4}$.
 Now we can write the following:
 \begin{eqnarray}
  &&\Pr_{I}\left\{\min\limits_{1\leq t \leq T^*} \left(K_{Q}(t) -t\left(\frac{1}{2}-2\epsilon\right)\right) \leq -\sqrt{T^*}\right\} \nonumber\\
    & \leq & \Pr_{I}\left\{\max\limits_{1\leq t \leq T^*} \left|K_{Q}(t) -t\left(\frac{1}{2}-2\epsilon\right)\right|  \geq  \sqrt{T^*}\right\}\nonumber\\
    & \leq & \frac{Var[K_{Q}(T^*)]}{T^*} < \frac{1}{4},
\end{eqnarray}
 wherein we have used Kolmogorov's inequality.
\end{proof}

\begin{lemma}\label{lem:lb1}
 Let $I \in \I_{k-1}$ and $Q \in \I_{m-k+1}$ such that $Q \subseteq \bar{I}$,
 and let $W$ be some fixed sequence of rewards
 obtained by a single run of algorithm $\mathcal{L}$ on $\mathcal{B}^{I}$ such
 that $T_{Q} \leq T^*$  and $K_{Q} \geq \frac{T_{Q}}{2} - \Delta$, then:
 \begin{equation}
  \Pr_{I \cup Q}\{W\} > \Pr_{I}\{W\}\cdot \exp(-32\epsilon\Delta).
 \end{equation}
\end{lemma}
\begin{proof}
Recall the fact that  all the arms in $Q$ have the same mean. Hence, if chosen
one at each trial (following any strategy), the expected reward at each trial
remains the same. Hence the probability of getting a given reward sequence generated from $Q$
is independent of the sampling strategy.
% by following any sequential strategy will be the same.
%  as the probability of generating that reward sequence by pulling any fixed arm in $Q$.
Again as the arms in
$Q$ have higher mean in $\mathcal{B}^{Q}$, the probability of getting
the sequence (of rewards) decreases monotonically as the 1-rewards for $\mathcal{B}^{I_0}$ become fewer.
So we get
\begin{align}
& \Pr_{I \cup Q}\{W\}  =  \Pr_{I}\{W\} \frac{\left(\frac{1}{2} + 2\epsilon\right)^{K_{Q}} \left(\frac{1}{2} - 2\epsilon\right)^{T_{Q}-K_{Q}}}{\left(\frac{1}{2} - 2\epsilon\right)^{K_{Q}} \left(\frac{1}{2} + 2\epsilon\right)^{T_{Q}-K_{Q}}} \nonumber\\
& \geq \Pr_{I}\{W\} \frac{\left(\frac{1}{2} + 2\epsilon\right)^{\left(\frac{T_{Q}}{2}-\Delta\right)} \left(\frac{1}{2} - 2\epsilon\right)^{\left(\frac{T_{Q}}{2}+\Delta\right)}}{\left(\frac{1}{2} - 2\epsilon\right)^{\left(\frac{T_{Q}}{2}-\Delta\right)} \left(\frac{1}{2} + 2\epsilon\right)^{\left(\frac{T_{Q}}{2}+\Delta\right)}} \nonumber\\
& = \Pr_{I}\{W\}\cdot \left(\frac{\frac{1}{2} - 2\epsilon}{\frac{1}{2} + 2\epsilon}\right)^{2\Delta} \nonumber\\
& >  \Pr_{I}\{W\}\cdot \exp(-32\epsilon\Delta) \left[\text{ for } 0 < \epsilon \leq \frac{1}{\sqrt{32}}\right]\nonumber.
\end{align}
\end{proof}

\begin{lemma}\label{lem:boundallreward}
  If \eqref{eq:lbusage} holds for an $I \in \I_{k-1}$ and $Q \in \I_{m-k+1}$ such that $Q \subseteq \bar{I}$,
  and if $\mathcal{W}$ is the set of all
 possible reward sequences $W$, obtained by algorithm $\mathcal{L}$ on $\mathcal{B}^{I}$, then $\Pr_{I \cup Q}\{\mathcal{W}\} > \left(\Pr_{I}\left\{\mathcal{W}\right\}-\frac{1}{2}\right)\cdot 4\delta.$
%   \begin{equation*}
%     \Pr_{I \cup Q}\{\mathcal{W}\} > \left(\Pr_{I}\left\{\mathcal{W}\right\}-\frac{1}{2}\right)\cdot 4\delta.
%   \end{equation*}
   In particular,
   \begin{equation}\label{eq:err2delta}
    \Pr_{I \cup Q}\{S_\mathcal{L} \subseteq I_0 \cup I\} > \frac{\delta}{\binom{m}{m-k+1}}.
   \end{equation}

\end{lemma}
\begin{proof}
Let for some fixed sequence (of rewards) $W$, $T_{Q}^W$ and $ K_{Q}^W $ respectively
denote the total number of samples received by the arms in $Q$ and the total number of $1$-rewards obtained before the algorithm $\mathcal{L}$ stopped.
Then:
\begin{align*}
 &\Pr_{I \cup Q}\{W\}  = \Pr_{I \cup Q}(W : W \in \mathcal{W}) \nonumber\\
& \geq  \Pr_{I \cup Q}\left\{W : W \in \mathcal{W} \bigwedge T_{Q}^W \leq T^* \bigwedge  K_{Q}^W \geq \frac{T_{Q}^W}{2} - \Delta\right\}\\
& >  \Pr_{I}\left\{W : W \in \mathcal{W} \bigwedge T_{Q}^W \leq T^* \bigwedge  K_{Q}^W \geq \frac{T_{Q}^W}{2} - \Delta\right\}\cdot \exp(-32\epsilon\Delta)\\
& \geq \left(\Pr_{I}\left\{W : W \in \mathcal{W} \bigwedge T_{Q}^W \leq T^*\right\}-\frac{1}{4}\right)\cdot \exp(-32\epsilon\Delta)\\
& \geq  \left(\Pr_{I}\left\{\mathcal{W}\right\}-\frac{1}{2}\right)\cdot \frac{4\delta}{\binom{m}{m-k+1}}\; \text{ for } C = \frac{1}{18375},\; \delta < \frac{e^{-1}}{4} .\\
\end{align*}
 In the above, the $3^\text{rd}$, $4^\text{th}$ and the last step are obtained using Lemma~\ref{lem:lb1}, Lemma~\ref{lem:boundreward} and Equation~\eqref{eq:lbusage}  respectively.
The inequality~\eqref{eq:err2delta} is obtained by using inequality~\eqref{eq:contra},
as $\Pr_{I}\{S_\mathcal{L} \in I_0 \} > 1 - \delta \geq 1 - \frac{e^{-1}}{4} > \frac{3}{4}$.
\end{proof}
\subsubsection{Summing Over $\I_{k-1}$ and $\I_m$}
Now, we sum up the probability of errors across all the instances in $\I_{k-1}$ and $\I_m$.
If the Assumption~\ref{asmp:contra} is true, using the pigeon-hole principle we show that
there exists some instance for which the mistake probability is greater than $\delta$.

\begin{align*}
 & \sum_{J \in \I_m} \Pr_J\{S_\mathcal{L} \nsubseteq J\} \\
 & \geq \sum_{J \in \I_m} \sum_{\substack{J' \subset J\\ : |J'| = m-k+1}} \Pr_J\{S_\mathcal{L} \subseteq \{J \setminus J'\} \cup I_0\} \\
 & \geq \sum_{J \in \I_m} \sum_{\substack{J' \subset J\\ : |J'| = m-k+1}} \Pr_J\{\exists a \in I_0: S_\mathcal{L} = \{J \setminus J'\} \cup\{a\}\} \\
 & = \sum_{J \in \I_m} \sum_{\substack{J' \subset J\\ : |J'| = m-k+1}} \sum_{I\in\I_{k-1}} \idop[I \cup J'= J] \cdot \Pr_J\{S_\mathcal{L} \subseteq I \cup I_0\}\\
 & = \sum_{J \in \I_m} \sum_{\substack{J' \subset \mathcal{A}\setminus I_0\\ : |J'| = m-k+1}} \sum_{I\in\I_{k-1}} \idop[I \cup J'= J] \cdot \Pr_J\{S_\mathcal{L} \subseteq I \cup I_0\}\\
 & = \sum_{J \in \I_m}  \sum_{I\in\I_{k-1}} \sum_{\substack{J' \subset \mathcal{A}\setminus I_0\\ : |J'| = m-k+1}}\idop[I \cup J'= J] \cdot \Pr_J\{S_\mathcal{L} \subseteq I \cup I_0\}\\
 & = \sum_{I\in\I_{k-1}} \sum_{J \in \I_m} \sum_{\substack{J' \subset \bar{I}\\ : |J'| = m-k+1}}\idop[I \cup J'= J] \cdot \Pr_J\{S_\mathcal{L} \subseteq I \cup I_0\}\\
 & = \sum_{I\in\I_{k-1}} \sum_{\substack{J' \subset \bar{I}\\ : |J'| = m-k+1}} \sum_{J \in \I_m} \idop[I \cup J'= J] \cdot \Pr_J\{S_\mathcal{L} \subseteq I \cup I_0\}\\
 & = \sum_{I\in\I_{k-1}} \sum_{\substack{J' \subset \bar{I}\\ : |J'| = m-k+1}} \Pr_{I\cup J'}\{S_\mathcal{L} \subseteq I \cup I_0\}
\end{align*}

Recall that $\forall I \in \I_{k-1}$ there exists a set $Q \subset \mathcal{A}\setminus\{I\cup I_0\}: |Q| = (m-k+1)$,
such that $T_Q < T^*$. Therefore,
\begin{align*}
 & \sum_{J \in \I_m} \Pr_J\{S_\mathcal{L} \nsubseteq J\}\\
 & \geq \sum_{I\in\I_{k-1}} \sum_{\substack{J' \subset \bar{I}\\ : |J'| = m-k+1}} \Pr_{I\cup J'}\{S_\mathcal{L} \subseteq I \cup I_0\}\\
 & > \sum_{I\in\I_{k-1}} \sum_{\substack{J' \subset \bar{I}\\ : |J'| = m-k+1}} \frac{\delta}{\binom{m}{m-k+1}}\\
 & \geq \sum_{I\in\I_{k-1}} \binom{n-m}{m-k+1} \cdot \frac{\delta}{\binom{m}{m-k+1}}\\
 & \geq \binom{n-(m-k+1)}{k-1} \cdot \binom{n-m}{m-k+1} \cdot \frac{\delta}{\binom{m}{m-k+1}}\\
 & = \binom{n-(m+k-1)}{m} \delta\\
 & = |\I_m| \delta.
\end{align*}

Hence, we get a  contradiction to Assumption~\ref{asmp:contra}, thereby proving the theorem.
% for $m \leq 2k -2$ and $m \geq 2k$.

% \subsection{Case $m = 2k -1 < \frac{n}{2}$:}

% Let $m_0 = 2k-1$, and let $N_0$ be the worst case sample complexity for $m = m_0$.
% We notice that the identification of $k$ arms from the best
% $m$ arms becomes harder as $m$ decreases.
% Therefore, $N_0$ is not lesser than the minimum number of samples required for $m = m_0 + 1$, and from the previous case we can write
% $ N_0 \geq \frac{1}{18375}\frac{1}{\epsilon^2}. \frac{n}{(m_0+1)-k+1}\ln\left(\frac{\binom{m}{m-k+1}}{4\delta}\right)
%   \geq \frac{1}{36750}\frac{1}{\epsilon^2}. \frac{n}{m_0-k+1}\ln\left(\frac{\binom{m}{m-k+1}}{4\delta}\right).$
% % Therefore, 
% % $$
% % N_0 \geq \frac{1}{18375}\frac{1}{\epsilon^2}. \frac{n}{(m_0+1)-k+1}\ln\left(\frac{\binom{m}{m-k+1}}{4\delta}\right) \geq \frac{1}{36750}\frac{1}{\epsilon^2}. \frac{n}{m_0-k+1}\ln\left(\frac{\binom{m}{m-k+1}}{4\delta}\right).
% % $$
% Hence, Theorem~\ref{thm:lbmainthm} is proved.
\pagebreak
\onecolumn
\section{Analysis of \GLUCB}
\label{app:adaptive}

Let at time $t$, $\hatp_a^t$ be the empirical mean of the arm $a \in \mathcal{A}$,
and $u_a^t$ be the number of times the arm $a$ has been pulled
until (and excluding) time $t$. For a given $\delta \in (0,1]$, we define
$\beta(u_a^t, t, \delta) = \sqrt{\frac{1}{2u_a^t}\ln\frac{k_1 n t^4}{\delta}}$, where $k_1=5/4$.
We define upper and lower confidence bound on the estimate of the true
mean of arm $a \in \mathcal{A}$ as $ucb(a,t) = \hatp_a + \beta(u_a^t, t, \delta)$,
and $lcb(a,t) = \hatp_a - \beta(u_a^t, t, \delta)$ respectively.

% \begin{algorithm}[]
% \label{alg:glucb}
% \caption{\textsc{GLUCB}:$(\epsilon, k, m)$-optimal subset selection}
% % \dontprintsemicolon % Some LaTeX compilers require you to use
% % instead
%  \KwIn{$\mathcal{A}$ (\st $|\mathcal{A}| = n$), $k, m, \epsilon, \delta$}
%  \KwOut{$(\epsilon, k, m)$-optimal subset of $\mathcal{A}$}
%  Pull each arm $a  \in \mathcal{A}$ for once. Set $t = n$
%  \Do{$ ucb({l_*^t}, t+1) - lcb({h_*^t}, t+1) > \epsilon $} { \label{ln:stpkoutofm}
% 	 $t = t + 1$
% 	 $A_1^t = \{a : a' \in \mathcal{A}, \hatp_a = \hatp_{a'}\}$ \st $|A_1^t| = k$
% 	 $A_3^t = \{a : a' \in \mathcal{A}, \hatp_a = \hatp_{a'}\}$ \st $|A_3^t| = n-m$
% 	 $A_2^t = \{\mathcal{A} \setminus (A_1^t \cup A_3^t)\}$
% 	 $h_*^t = \arg \max_{\{a \in A_1^t\}} lcb(a,t)$
% 	 $m_*^t = \arg \min_{\{a \in A_2^t\}} u_a^{t}$
% 	 $l_*^t = \arg \max_{\{a \in A_3^t\}} ucb(a,t)$
% 	 pull  $h_*^t,  m_*^t, l_*^t$
%  }
%  \Return $A_1^t$
% \end{algorithm}

% %%%%%%%%% Expected Sample Complexity of \GLUCB %%%%
% \thmscglucb*
% %%%%%%%%%%%%%%
To analyse the sample complexity, first we define some events, at least
one of which must occur if the algorithm does not stop at the round $t$.

\begin{definition}{(\textsc{Probable Events})}
Let $a, b \in \mathcal{A}$, such that $\mu_a > \mu_b$. During the
run of the algorithm, any of the following five events may occur:\\
i) The empirical mean of an arm may falls outside the upper or the lower
confidence bound. We define it as:
$$CROSS_a^t \defeq \{ucb(a,t) < \mu_a \vee lcb(a,t) > \mu_a\}.$$

ii) The empirical mean of arm $a$ may be lesser than that of arm $b$; we definite as:
$$ErrA(a,b,t) \defeq \{\hatp_a^t < \hatp_b^t\}.$$

iii) The lower and upper confidence bounds of arm $a$ may fall below those of arm $b$; we
define them as:
\begin{align*}
 & ErrL(a,b,t) \defeq \{lcb(a,t) < lcb(b,t)\},\\
 & ErrU(a,b,t) \defeq \{ucb(a,t) < ucb(b,t)\}.
\end{align*}

iv) If an arm's confidence bounds are above a certain radius (say $d$), we define that event as
\begin{equation*}
 NEEDY_a^t(d) \defeq \left\{\{lcb(a,t) < \mu_a -d\} \vee \{ucb(a,t) > \mu_a +d\}\right\}.
\end{equation*}
\end{definition}

We show that any arm $a$, if sampled sufficiently, that is $u_a^t \geq u^*(a,t)$, 
then occurrence of any of the \textsc{Probable Events} imply occurrence of $CROSS_a^t$.
First we show that if  $CROSS_a^t$ does not occur for any $a \in \A$, then occurrence
of any one of the \textsc{Probable Events} implies the occurrence of $NEEDY_a^t(\cdot)$
or $NEEDY_b^t(\cdot)$.

%%%%%%% Reducing Events To $NEEDY_a^t$ %%%%%%%%
\begin{restatable}{lemma}{lemErrALUN}[Expressing \textsc{Probable Events} in terms of $NEEDY_a^t$ and $CROSS_a^t$]
\label{lem:ErrALUN}
To prove that $\{\neg  CROSS_a^t \wedge \neg CROSS_b^t\} \wedge \{ErrA(a,b,t) \vee ErrU(a,b,t) \vee ErrL(a,b,t)\} \implies \{NEEDY_a^t\left(\frac{\Delta_{ab}}{2}\right) \vee NEEDY_b^t\left(\frac{\Delta_{ab}}{2}\right) \}$.
\end{restatable}
%%%%%%%%%%%%%%
\begin{proof}
$\mathbf{ErrA(a,b,t)}$: To prove that $\neg \{CROSS_a^t \vee CROSS_b^t\} \wedge ErrA(a,b,t) \implies  NEEDY_a^t\left(\frac{\Delta_{ab}}{2}\right) \vee NEEDY_b^t\left(\frac{\Delta_{ab}}{2}\right)$.
      \begin{align*}
      & ErrA(a,b,t) \implies \hat{p}_a^t < \hat{p}_b^t \\
      & \implies \hat{p}_a^t - (p_a - \beta(u_a^t,t,\delta)) < \hat{p}_b^t - (p_b + \beta(u_b^t,t,\delta) +\\
      & (\beta(u_a^t,t,\delta) + \beta(u_b^t,t,\delta)) - \Delta_{ab}/2)\\
      &\implies  NEEDY_a^t\left(\frac{\Delta_{ab}}{2}\right) \vee NEEDY_b^t\left(\frac{\Delta_{ab}}{2}\right). 
      \end{align*}

$\mathbf{ErrU(a,b,t)}$: To prove that $\neg \{CROSS_a^t \vee CROSS_b^t\} \wedge ErrU(a,b,t) \implies NEEDY_b^t\left(\frac{\Delta_{ab}}{2}\right)$.\\
Assuming $\neg CROSS_a^t \wedge \neg CROSS_b^t$ we get
\begin{align*}
& ErrU(a,b,t) \implies \{ucb(b,t) > ucb(a,t)\}\\
& \implies \{\hat{p}_b^t + \beta(u_b^t,t,\delta)> \hat{p}_a^t + \beta(u_a^t,t,\delta)\}\\
& \implies \{\hat{p}_b^t > \mu_b + \beta(u_b^t,t,\delta)\} \vee \{\hat{p}_a^t < \mu_a - \beta(u_a^t,t,\delta)\}  \vee\\ 
& \hspace{28pt} \{2\beta(u_b^t,t,\delta) > \Delta_{ab}\}\\
& \implies NEEDY_b^t\left(\frac{\Delta_{ab}}{2}\right).
\end{align*}

$\mathbf{ErrL(a,b,t)}$: To prove that $\neg \{CROSS_a^t \vee CROSS_b^t\} \wedge ErrL(a,b,t) \implies NEEDY_a^t\left(\frac{\Delta_{ab}}{2}\right)$.\\
Assuming $\neg CROSS_a^t \wedge \neg CROSS_b^t$ we get
\begin{align*}
& ErrL(a,b,t) \implies \{lcb(b,t) > lcb(a,t)\}\\
& \implies \{\hat{p}_b^t - \beta(u_b^t,t,\delta)> \hat{p}_a^t - \beta(u_a^t,t,\delta)\}\\
& \implies \{\hat{p}_b^t > \mu_b + \beta(u_b^t,t,\delta)\} \vee \{\hat{p}_a^t < \mu_a - \beta(u_a^t,t,\delta)\}  \vee\\ 
& \hspace{28pt} \{2\beta(u_a^t,t,\delta) > \Delta_{ab}\}\\
& \implies NEEDY_a^t\left(\frac{\Delta_{ab}}{2}\right).
\end{align*}
\end{proof}
We show that given a threshold $d$, if an arm $a$ is sufficiently sampled, such that $\beta(u_a^t, t, \delta) \leq \frac{d}{2}$, then $NEEDY_a^t$ infers $CROSS_a^t$.

%%%%%% NEEDY TO CROSS %%%%%%%%%%%
\begin{restatable}{lemma}{lemneedycross}
 \label{lem:needycross}
  For any $a \in \A$, $\{NEEDY_a^t(d)|\beta(u_a^t, t, \delta) < d/2\} \implies CROSS_a^t$.
\end{restatable}
%%%%%%%%%%%%%%%%

\begin{proof}
 First, we show that $\{lcb(a,t) < \mu_a -d | \beta(u_a^t, t, \delta) < d/2\} \implies CROSS_a^t$,
 \begin{align}
  & \{lcb(a,t) < \mu_a -d | \beta(u_a^t, t, \delta) < d/2\} \nonumber\\
  & \implies \{\hatp_a^t - \beta(u_a^t, t, \delta) < \mu_a -d | \beta(u_a^t, t, \delta) < d/2\} \nonumber\\
  & \implies \{\hatp_a^t < \mu_a -d +\beta(u_a^t, t, \delta) | \beta(u_a^t, t, \delta) < d/2\} \nonumber\\
  & \implies \{\hatp_a^t < \mu_a -d/2 | \beta(u_a^t, t, \delta) < d/2\} \nonumber\\
  & \implies CROSS_a^t.
 \end{align}
 The other part follows the similar way.
\end{proof}

By the very definition of confidence bound, at any round $t$, the probability that
the empirical mean of an arm will lie outside it, is very low. In other words, the
probability of occurrence $CROSS_a^t$ is very low for all $t$ and $a \in \A$.

%%%%%%%%%% Upper bounding the probability of $CROSS_a^t$ %%%%%%%%%%%%%%%%%
\begin{restatable}{lemma}{lemcross}[Upper bounding the probability of $CROSS_a^t$]
 \label{lem:cross}
 $\forall a \in \mathcal{A}$ and $\forall t \geq 0$, $\Pr\{{CROSS_a^t}\}  \leq  \frac{\delta}{knt^4}$. Hence,
 $P\left[\exists t \geq 0  \wedge \exists a \in \mathcal{A} : {CROSS_a^t} | u_a^t \geq 0  \right] \leq  \frac{\delta}{k_1 t^3}.$
\end{restatable}
%%%%%%%%%%%%
\begin{proof}
$\Pr\{{CROSS_a^t}\}$ is upper bounded by using Hoeffding's inequality, and the next statement
gets proved by taking union bound over all arms and $t$.
\end{proof}
Now, recalling the definition of $h_*^t$, and $l_*^t$ from Algorithm~\ref{alg:glucb},
we present the key logic underlying the analysis of \GLUCB. The idea is to show that
if the algorithm has not stopped, then one of those \textsc{Probable Events} must have
occurred. Then using Lemma~\ref{lem:ErrALUN},
 and Lemma~\ref{lem:needycross}, Lemma~\ref{lem:cross}
we show that beyond a certain number of rounds, the probability that \GLUCB
will continue is sufficiently small.
Lastly, using the argument based on pigeon-hole principle, similar to
Lemma 5 of 
\citet{bib:shivaramphdthesis}, we establish the upper bound on the 
sample complexity. Below we present the core logic that shows, until the algorithm stops one of the
\textsc{Probable Events} must occur.

%%%%%%%%%%%%%%%%%%%%%%
%      Case 1
%%%%%%%%%%%%%%%%%%%%%%
\setcounter{logicase}{0}
\begin{logicase}[H]
\begin{algorithmic}
  \IF {$\exists b_3 \in A_1^t \cap B_3$}{
    \STATE Then $ErrL(h_*^t, b_3, t)$ has occurred.
  }\ELSE {
    \STATE $\exists b_3 \in A_2^t \cap B_3$
    \STATE Then $ErrA(h_*^t, b_3, t)$  has occurred.
%     Then $CROSS_{h_*^t}^t \vee CROSS_{b_3}^t$ has occurred
  }\ENDIF
\end{algorithmic}
\label{case:1}
\caption{$h_*^t \in B_1 \wedge l_*^t \in B_1$}
\end{logicase}

%%%%%%%%%%%%%%%%%%%%%%
%      Case 2
%%%%%%%%%%%%%%%%%%%%%%
\begin{logicase}[H]
\begin{algorithmic}
  \IF {$\exists b_3 \in A_1^t \cap B_3$}{
      \STATE Then $ErrL(h_*^t, b_3^t, t)$ has occurred.
  }\ELSE {
    \STATE $\exists b_3 \in A_2^t \cap B_3$.
    \IF {$\Delta_{h_*^t l_*^t} \geq \frac{\Delta_{h_*^t}}{2}$} {
      \STATE Then $NEEDY_{h_*^t}^t(\Delta_{h_*^t}/4) \vee NEEDY_{l_*^t}^t(\Delta_{h_*^t}/4)$ has occurred.
    }\ELSE  {
      \STATE Then $ErrL(l_*^t, b_3^t, t)$ has occurred.
    }\ENDIF
  }\ENDIF
\label{case:2}
\caption{$h_*^t \in B_1 \wedge l_*^t \in B_2$}
\end{algorithmic}
\end{logicase}

%%%%%%%%%%%%%%%%%%%%%%
%      Case 3
%%%%%%%%%%%%%%%%%%%%%%
\begin{logicase}[H]
\begin{algorithmic}
   \STATE Then $NEEDY_{h_*^t}^t(\Delta_{h_*^t}/4) \vee NEEDY_{l_*^t}^t(\Delta_{l_*^t}/4)$ has occurred.
\end{algorithmic}
\label{case:3}
\caption{$h_*^t \in B_1 \wedge l_*^t \in B_3$}
\end{logicase}

%%%%%%%%%%%%%%%%%%%%%%
%      Case 4
%%%%%%%%%%%%%%%%%%%%%%
\begin{logicase}[H]
\begin{algorithmic}
  \IF {$\Delta_{h_*^t l_*^t} \geq \frac{\Delta_{h_*^t}}{2}$}{
    \STATE Then $ErrA(l_*^t, h_*^t, t)$  has occurred.
%     Then $CROSS_{h_*^t}^t \vee CROSS_{l_*^t}^t$ has occurred
  }\ELSE {
    \IF {$\exists b_3 \in A_1^t \cap B_3$}{
      \STATE Then $ErrL(h_*^t, b_3^t, t)$ has occurred.
    }\ELSE {
      \STATE $\exists b_3 \in A_2^t \cap B_3$
      \STATE $\therefore ErrA(l_*^t, b_3, t)$  has occurred.
%       $\therefore CROSS_{l_*^t}^t \vee CROSS_{b_3}^t$ has occurred.
    }\ENDIF
  }\ENDIF
\end{algorithmic}
\label{case:4}
\caption{$h_*^t \in B_2 \wedge l_*^t \in B_1$}
\end{logicase}

%%%%%%%%%%%%%%%%%%%%%%
%      Case 5.A
%%%%%%%%%%%%%%%%%%%%%%

\begin{logicase}[H]
\begin{algorithmic}
 \STATE Here, $\exists b_1 \in (A_2^t \cup A_3^t)\cap B_1$ and $\exists b_3 \in (A_1^t \cup A_2^t)\cap B_3$
  \IF {$|\Delta_{h_*^t l_*^t}| < \Delta_{h_*^t}/2$}{
    \IF {$\Delta_{b_1 h_*^t} > \Delta_{b_1}/4$}{
      \IF {$b_1 \in A_2^t )\cap B_1$}{
	\STATE $ErrA(b_1, h_*^t, t)$
      }\ELSE {
	\STATE $b_1 \in A_3^t \cap B_1$
	\STATE $ErrU(b_1, l_*^t, t)$ has occurred.
      }\ENDIF
    }\ELSE {
      \STATE $\Delta_{b_1 h_*^t} \leq \Delta_{b_1}/4$  and hence $\Delta_{l_*^t b_3} \geq \Delta_{l_*^t}/4$
      \IF {$b_3 \in A_2^t \cap B_3$}{
	\STATE $ErrA(l_*^t, b_3, t)$ has occurred.  
      }\ELSE {
	\STATE $b_3 \in A_1^t \cap B_3$
	\STATE $ErrL(h_*^t, b_3, t)$ has occurred.
      }\ENDIF
    }\ENDIF
  }\ELSE {
    \STATE $|\Delta_{h_*^t l_*^t}| > \Delta_{h_*^t}/2$
    \STATE $NEEDY_{h_*^t}^t (\Delta_{h_*^t}/4) \vee NEEDY_{l_*^t}^t (\Delta_{h_*^t}/4)$ has occurred.
  }\ENDIF
\end{algorithmic}
 \label{case:5.A}
\caption{$h_*^t \in B_2 \wedge l_*^t \in B_2$ and $\Delta_{h_*^t l_*^t} > 0$}
\end{logicase}

%%%%%%%%%%%%%%%%%%%%%%
%      Case 5.B
%%%%%%%%%%%%%%%%%%%%%%
\setcounter{logicase}{4}
\begin{logicase}[H]
\begin{algorithmic}
  \STATE Here, $\exists b_1 \in (A_2^t \cup A_3^t)\cap B_1$ and $\exists b_3 \in (A_1^t \cup A_2^t)\cap B_3$
  \IF {$|\Delta_{h_*^t l_*^t}| < \Delta_{h_*^t}/2$}{
    \IF {$\Delta_{b_1 l_*^t} > \Delta_{b_1}/4$}{
      \IF {$b_1 \in A_2^t \cap B_1$}{
	\STATE $ErrA(b_1, h_*^t, t)$ has occurred.
      }\ELSE {
	\STATE $b_1 \in A_3^t \cap B_1$
	\STATE $ErrU(b_1, l_*^t, t)$ has occurred.
      }\ENDIF
    }\ELSE {
      \STATE $\Delta_{b_1 l_*^t} \leq \Delta_{b_1}/4$ and hence $\Delta_{h_*^t b_3} \geq \Delta_{h_*^t}/4$
      \IF {$b_3 \in A_2^t \cap B_3$}{
	\STATE $ErrA(l_*^t, b_3, t)$ has occurred.  
      }\ELSE {
	\STATE $b_3 \in A_1^t \cap B_3$
	\STATE $ErrL(h_*^t, b_3, t)$ has occurred.
      }\ENDIF
    }\ENDIF
  }\ELSE {
    \STATE $|\Delta_{h_*^t l_*^t}| > \Delta_{h_*^t}/2$
    \STATE $NEEDY_{h_*^t}^t (\Delta_{h_*^t}/4) \vee NEEDY_{l_*^t}^t (\Delta_{h_*^t}/4)$ has occurred.
  }\ENDIF
\end{algorithmic}
  \label{case:5.B}
\caption{(continued) $h_*^t \in B_2 \wedge l_*^t \in B_2$ and $\Delta_{h_*^t l_*^t} \leq 0$} %. Similar to~\ref{case:5.A}
\end{logicase}

%%%%%%%%%%%%%%%%%%%%%%
%      Case 6
%%%%%%%%%%%%%%%%%%%%%%
\begin{logicase}[H]
\begin{algorithmic}
  \IF {$\Delta_{h_*^t l_*^t} \geq \frac{\Delta_{l_*^t}}{2}$}{
    \STATE Then $NEEDY_{h_*^t}^t(\Delta/4) \vee NEEDY_{l_*^t}^t(\Delta_{l_*^t}/4)$ has occurred.
  }\ELSE {
    \STATE $\Delta_{h_*^t l_*^t} < \frac{\Delta_{l_*^t}}{2}$
    \STATE $\therefore \forall b_1 \in \{A_2^t \cup A_3^t\} \cap B_1$, $\Delta_{b_1 h_*^t} > \frac{\Delta_{b_1}}{2}$.
    \IF {$\exists b_1 \in A_2^t \cap B_1$}{
      \STATE $ErrA(b_1, h_*^t, t)$ has occurred.
%       $\therefore CROSS_{h_*^t}^t \vee CROSS_{b_1}^t$ has occurred.
    }\ELSE {
      \STATE $\exists b_1 \in A_3^t \cap B_1$.
      \STATE Then $ErrU(b_1^t, l_*^t,t)$ has occurred.
%       OR
%       $CROSS_{l_*^t}^t \vee CROSS_{b_1}^t$ has occurred.
    }\ENDIF
  }\ENDIF
\end{algorithmic}
\label{case:6}
\caption{$h_*^t \in B_2 \wedge l_*^t \in B_3$}
\end{logicase}

%%%%%%%%%%%%%%%%%%%%%%
%      Case 7
%%%%%%%%%%%%%%%%%%%%%%
\begin{logicase}[H]
\begin{algorithmic}
  \STATE $\therefore ErrA(l_*^t, h_*^t, t)$ has occurred.
%   $CROSS_{h_*^t}^t \vee CROSS_{l_*^t}^t$ has occurred.
\end{algorithmic}
\label{case:7}
\caption{$h_*^t \in B_3 \wedge l_*^t \in B_1$}
\end{logicase}

%%%%%%%%%%%%%%%%%%%%%%
%      Case 8
%%%%%%%%%%%%%%%%%%%%%%

\begin{logicase}[H]
\begin{algorithmic}
  \IF {$\Delta_{h_*^t l_*^t} \geq \frac{\Delta_{h_*^t}}{2}$}{
    \STATE $ErrA(l_*^t, h_*^t, t)$ has occurred.
%     Then $CROSS_{h_*^t}^t \vee CROSS_{l_*^t}^t$ has occurred.
  }\ELSE {
    \STATE $\Delta_{h_*^t l_*^t} < \frac{\Delta_{h_*^t}}{2}$
    \STATE $\therefore \forall b_1 \in \{A_2^t \cup A_3^t\} \cap B_1$, $\Delta_{b_1 l_*^t} > \frac{\Delta_{b_1}}{2}$.
    \IF {$\exists b_1 \in A_2^t \cap B_1$}{
      \STATE $ErrA(b_1, h_*^t, t)$ has occurred.
%       $\therefore CROSS_{h_*^t}^t \vee CROSS_{b_1}^t$ has occurred.
    }\ELSE {
      \STATE $\exists b_1 \in A_3^t \cap B_1$.
      \STATE $\therefore ErrU(b_1, l_*^t, t)$ has occurred.
    }\ENDIF
  }\ENDIF
\end{algorithmic}
\label{case:8}
\caption{$h_*^t \in B_3 \wedge l_*^t \in B_2$}
\end{logicase}

%%%%%%%%%%%%%%%%%%%%%%
%      Case 9
%%%%%%%%%%%%%%%%%%%%%%
\begin{logicase}[H]
\begin{algorithmic}
  \STATE $\exists b_1 \in \{A_2^t \cup A_3^t\} \cap B_1$ %, $\Delta_{1l_*^t} > \frac{\Delta}{2}$.
  \IF {$\exists b_1 \in A_2^t \cap B_1$} {
    \STATE $ErrA(b_1, h_*^t, t)$ has occurred.
  }\ELSE {
    \STATE $\exists b_1 \in A_3^t \cap B_1$
    \STATE $\therefore ErrA(b_1, l_*^t, t)$ has occurred.
  }\ENDIF
%   $\therefore {b_1}^t \vee CROSS_{h_*^t}^t \vee CROSS_{h_*^t}^t$ has occurred.
\end{algorithmic}
\label{case:9}
\caption{$h_*^t \in B_3 \wedge l_*^t \in B_3$}
\end{logicase}
%%%%%%%%%%%%%%%%%%%%%%%%%%%%%%%%%%%%%%%%%%%%%%%%%%%%%%%%%%%%%%%%%%%%

% \corgenubscglucb*
%%%%%%%%%%%%%%%%%%%%%%%%%%%%%%%%%%%%%%%%%%%%%%%%%%%%%%%%%%%%%%%%%%%%

\begin{lemma}[H]
\label{lem:valT}
 If $T = C H_\epsilon\ln\left(\frac{H_\epsilon}{\delta}\right)$, then
 for $C \geq 2732$, the following holds: $$T > 2 + 2 \sum\limits_{a \in \mathcal{A}} u^*(a,T).$$
\end{lemma}
\begin{proof}
This proof is taken from Appendix B.3 of \citet{bib:shivaramphdthesis}.
\begingroup
\allowdisplaybreaks
\begin{align*}
     &2 + 2 \sum_{a \in \mathcal{A}} u^*(a,T) =  2 + 64 \sum_{a \in \mathcal{A}} \bceil{\frac{1}{\max(\Delta_{a}, (\epsilon / 2))^2} \ln \frac{ knt^4}{\delta}}\\
    & \leq  2 + 64n + 64 H_\epsilon\ln \frac{ knT^4}{\delta} \\
    & =  2 + 64n + 64H_\epsilon\ln k + 64H_\epsilon \ln\frac{n}{\delta} + 256 H_\epsilon \ln T\\
    & <  (66 + 64\ln k )H_\epsilon + 64H_\epsilon\ln\frac{n}{\delta}  + 256 H_\epsilon \left[\ln C + \ln H_\epsilon + \ln\ln\frac{H_\epsilon}{\delta}\right]\\
    & <  (66 + 64\ln k )H_\epsilon + 64H_\epsilon\ln\frac{n}{\delta} + 256 H_\epsilon \left[\ln C + \ln H_\epsilon + \ln\ln\frac{H_\epsilon}{\delta}\right]\\
    & <  130H_\epsilon + 64H_\epsilon\ln\frac{n}{\delta} + 256 H_\epsilon \left[\ln C + \ln H_\epsilon + \ln\frac{H_\epsilon}{\delta}\right]\\
    & <  130 H_\epsilon + 64H_\epsilon\ln\frac{H_\epsilon}{\delta}  + 256 H_\epsilon \left[\ln C + 2\ln\frac{H_\epsilon}{\delta}\right]\\
    & < (706 + 256\ln C) H_\epsilon\ln\frac{H_\epsilon}{\delta} <  C H_\epsilon\ln\frac{H_\epsilon}{\delta}\phantom{=}\,\left[\text{For }  C \geq 2732\right].
\end{align*}
\endgroup
\end{proof}

\begin{lemma}\label{lem:t1star}
 Let $T^* = \bceil{2732H_\epsilon\ln\left(\frac{H_\epsilon}{\delta}\right)}$.
 For every $T > T_1^*$, the probability that the Algorithm~\ref{alg:glucb}
 has not terminated after $T$ rounds of sampling is at most $\frac{8\delta}{T^2}$.
\end{lemma}
\begin{proof}
 Letting $\bar{T} = \frac{T}{2}$ we define two events for $ \bar{T} \leq t \leq T-1$:
   $E^{(1)} \defeq \exists a \in \mathcal{A} : {CROSS_a^t}$ and $E^{(2)} \defeq \exists  NEEDY_{a}^t\left(\frac{\Delta_a}{4}\right)$.
  If the algorithm stops for $t<\bar{T}$, then there is nothing to prove. On the contrary, let the algorithm has not stopped after
  $t > \bar{T}$ and neither $E^{(1)}$ nor $E^{(2)}$ has occurred. 
  Letting $N_{rounds}$ be the the  required number of rounds beyond $\bar{T}$,
  we can upper bound it as:
 \begin{align*}
  &N_{rounds} = \sum\limits_{t=\bar{T}} \left\{\idop\left[NEEDY_{h_*^t}^t\left(\frac{\Delta_{h_*^t}}{4}\right) \vee NEEDY_{m_*^t}^t\left(\frac{\Delta_{m_*^t}}{4}\right)  \vee   NEEDY_{l_*^t}^t\left(\frac{\Delta_{l_*^t}}{4}\right)\right]\right\}\\
  & \leq  \sum_{\bar{T}}^{T-1} \sum_{a \in \mathcal{A}}\idop\left[a \in \{h_*^t, m_*^t,  l_*^t\}\wedge NEEDY_{a}^t\left(\frac{\Delta_a}{4}\right)\right]\\
  & =  \sum_{\bar{T}}^{T-1} \sum_{a \in \mathcal{A}}\idop[a \in \{h_*^t, m_*^t, l_*^t\}\wedge(u_a^t < u^*(a,t))]\\
  & \leq  \sum_{\bar{T}}^{T-1} \sum_{a \in \mathcal{A}}\idop[a \in \{h_*^t, m_*^t, l_*^t\} \wedge(u_a^t < u^*(a,t))]\\
  & \leq  \sum_{a \in \mathcal{A}} \sum_{\bar{T}}^{T-1}\idop[(a \in \{h_*^t, m_*^t, l_*^t\})\wedge(u_a^t < u^*(a,t))]\\
  & \leq  \sum_{a \in \mathcal{A}}u^*(a,t).
 \end{align*}
  Using Lemma~\ref{lem:valT}, $T \geq T^* \Rightarrow T > 2 + 2\sum_{a \in \mathcal{A}}u^*(a,t)$.
 Hence, if neither $E^{(1)}$ nor $E^{(2)}$ occurs then the
 algorithm runs for at most $\bar{T} + N_{rounds} \leq \ceil{T/2} + \sum_{a \in \mathcal{A}}16u^*(a,t) < T$ 
 number of rounds.

 The probability that the algorithm does not stop within $T$ rounds, is upper-bounded
 by $P[E^{(1)} \vee E^{(2)}]$. Applying Lemma~\ref{lem:needycross} and Lemma~\ref{lem:cross},
 \begin{align*}
  & P[E^{(1)} \vee E^{(2)}] \leq \sum_{t = \bar{T}}^{T-1} \left(\frac{\delta}{k_1 t^3} + \frac{\delta}{kt^4}\right) \leq \sum_{t = \bar{T}}^{T-1} \frac{\delta}{k_1 t^3}\left(1+\frac{2}{t}\right) \leq  \left(\frac{T}{2}\right)\frac{8\delta}{k_1 T^3}\left(1 + \frac{4}{T}\right) < \frac{8\delta}{T^2}. %[\because n \geq 2]\nonumber
 \end{align*}
\end{proof}

\thmscglucb*
Using Lemma~\ref{lem:valT}, and Lemma~\ref{lem:t1star}
 the expected sample complexity of the Algorithm~\ref{alg:glucb} can be upper bounded as
 \begin{align}
  & E[SC] \leq 2\left(T_1^* + \sum_{t=T_1^*}^\infty \frac{8\delta}{T^2}\right) \leq 5464\cdot\left(H_\epsilon\ln\left(\frac{H_\epsilon}{\delta}\right)\right) + 32.
 \end{align}

\pagebreak
\section{Proof of Theorem~\ref{thm:qpreducubonly}}
\label{sec:hardnessqp}
% It is interesting to note that solving \QP problem using order-optimal sample complexity is crucially dependent on the
% existence of an algorithm that can solve \QF within order-optimal sample complexity for solving
% \QF. We start with proving the upper bound.

% \subsection{Upper Bound}
% \label{subsec:hardnessqpub}

% In this section we show that if there exists an algorithm for solving any instance of \QF
% within order-optimal sample-complexity, then we can construct an algorithm that can solve
% any instance of \QP  within order-optimal sample-complexity.

% \begin{restatable}{theorem}{thmqpubreduc}
% \label{thm:qpreducub}
% Let $\gamma: \mathbb{Z}^+ \times \mathbb{Z}^+ \times [0,1] \times [0,1] \mapsto \mathbb{R}^+$.
% If there exists an algorithm \textsc{OptQF} that can solve any instance of \QF  given by $(\A, n, m, 1, \epsilon, \delta)$
% within a sample-complexity $O\left(\frac{n}{m\epsilon^2}\log\frac{1}{\delta} + \gamma(\cdot)\right)$, then there exists an
% algorithm \textsc{OptQP} that can solve any instance of \QF  given by $(\A, P_\A, \rho, \epsilon, \delta)$
% within a sample-complexity  $O\left(\frac{1}{\rho\epsilon^2}\log\frac{1}{\delta} + \gamma(\cdot)\right)$.
% \end{restatable}
% \paragraph{Proof of Theorem~\ref{thm:qpreducub}.}
% We prove Theorem~\ref{thm:qpreducub} by construction. Following, we present an  algorithm \textsc{OptQP}
% that follows the approach similar to \textsc{OptQP}.
% It solves any instance of \QP  by reducing it to a finite instance to pose it as \QFK, and
% then solves that \QF   using \textsc{OptQF} as the subroutine.
 
Algorithm~\ref{alg:tightqpqf} describes \textsc{OptQP}. It uses \PP~\cite{bib:arcsk2017} with  \textsc{ Median Elimination}  as the 
subroutine (inside \PP), to select an  $[\epsilon, \rho]$-optimal arm with confidence $1-\delta'$.
We have assumed $\delta' = 1/4$, in practice the one can choose any
sufficiently small value for it, which will merely affect the multiplicative constant in the upper bound.
\begin{algorithm}[]
\begin{algorithmic}
\REQUIRE { $\mathcal{A}, \epsilon, \delta$, and \textsc{OptQF}.}
\ENSURE {A single $[\epsilon, \rho]$-optimal arm}
 \STATE Set $\delta' = 1/4$, $u = \bceil{\frac{1}{2(0.5-\delta')^2} \cdot \log\frac{2}{\delta}} = \bceil{8 \log\frac{2}{\delta}}$.\;
 \STATE Run $u$ copies of $\mathcal{P}_2(\A, \rho, \epsilon/2, \delta')$ and form set $S$ with the output arms.\;
 \STATE Return the output from \textsc{OptQF} $\left(S, u, \floor{\frac{u}{2}}, 1, \frac{\epsilon}{2}, \frac{\delta}{2}\right)$.
\end{algorithmic}
\caption{\textsc{OptQP}}
\label{alg:tightqpqf}
\end{algorithm}

 \begin{restatable}{theorem}{thmppp}[Correctness and Sample Complexity of \textsc{OptQP}]
 \label{thm:ppp}
 If \textsc{OptQF} exists, then
 \textsc{OptQP}  solves \QP, within the sample complexity $\Theta\left(\frac{1}{\rho\epsilon^2}\log\frac{1}{\delta}+ \gamma(\cdot)\right)$.
 \end{restatable}
 \begin{proof}
 First we prove the correctness and then upper bound the sample complexity.
  \paragraph{Correctness.}  First we notice that each copy of $\mathcal{P}_2$ outputs an $[\epsilon/2, \rho]$-optimal arm
 with probability at least $1-\delta'$. Also, \textsc{OptQF} outputs an
 $[\epsilon/2, \rho]$-optimal arm with probability $1-\delta$.
 Let, $\hat{X}$ be the fraction of sub-optimal arms in $S$. Then $\Pr\{\hat{X} \geq \frac{1}{2}\}$ $= \Pr\{\hat{X} - \delta' \geq \frac{1}{4}\}$
  $\leq \exp(-2\cdot(\frac{1}{4})^2\cdot u) = \exp(-2\cdot\frac{1}{16}\cdot 8\log\frac{2}{\delta}) < \frac{\delta}{2}$. On the other hand, the mistake probability of \textsc{OptQF} is upper bounded by $\delta/2$. Therefore, by taking union bound, we get the 
  mistake probability is upper bounded by $\delta$. Also, the mean of the output arm is not
  less than $\frac{\epsilon}{2} + \frac{\epsilon}{2} = \epsilon$ from the $(1-\rho)$-th
  quantile.
  
  \paragraph{Sample complexity.} First we note that, for some appropriate constant $C$,
  the sample complexity (SC) of each of the $u$ copies of $\mathcal{P}_2$ is $\frac{C}{\rho(\epsilon/2)^2}\left(\log\frac{2}{\delta'}\right)^2 \in O\left(\frac{1}{\rho\epsilon^2}\right)$.
  Hence, SC of all the $u$ copies $\mathcal{P}_2$ together is upper bounded by $\frac{C_1\cdot u}{\rho\epsilon^2}$, for some constant $C_1$.
  Also, for some constant $C_2$, the sample complexity of \textsc{OptQF} is upper bounded by $C_2 \left(\frac{u}{(u/2) (\epsilon/2)^2}\log\frac{2}{\delta} + \gamma(\cdot)\right) = C_2 \left(\frac{8}{\epsilon^2}\log\frac{2}{\delta}+ \gamma(\cdot)\right)$.
  Now, adding the sample complexities, and substituting for $u$ we prove the bound.
 \end{proof}

\end{appendices}
\end{document}